%% file: 00-Main.tex
\documentclass{interact}

\usepackage[colorlinks=true, allcolors=blue]{hyperref}
\usepackage{rotating}
\usepackage{booktabs}
\usepackage{tikz}

\usepackage{graphicx}
\usepackage{lineno}
\usepackage{orcidlink}
\usepackage[style = apa, backend = biber]{biblatex}

\usepackage{xcolor}

\newcommand{\highlighting}[1]{%
  {\color{black}#1}%
}

\usepackage{multirow}
\theoremstyle{plain}

\theoremstyle{definition}

\theoremstyle{remark}

\begin{document}

\articletype{Research Paper}% Specify the article type or omit as appropriate

\title{ConvVitMamba: Efficient Multiscale Convolution, Transformer, and Mamba-Based Sequence \highlighting{modelling} for Hyperspectral Image Classification}
\author{
\name{Mohammed.~Q. Alkhatib {\orcidlink{0000-0003-4812-614X}}\textsuperscript{a}
\thanks{CONTACT Mohammed.~Q. Alkhatib. Email: mqalkhatib@ieee.org }}
\affil{\textsuperscript{a}College of Engineering and IT, University of Dubai, Dubai 14143, United Arab Emirates}
}

\maketitle

\begin{abstract}
Hyperspectral image (HSI) classification remains a challenging task due to the high spectral dimensionality of the data, strong spectral redundancy, and limited availability of labeled samples. Although convolutional neural networks (CNNs) and Vision Transformers (ViTs) have demonstrated strong performance by exploiting spectral--spatial information and long-range dependencies, they often suffer from high computational complexity and large parameter counts, which limit their practical applicability. To address these limitations, a unified hybrid framework, termed ConvVitMamba, is proposed for efficient hyperspectral image classification. The proposed architecture integrates three complementary components within a single model: a multiscale convolutional feature extractor for capturing local spectral, spatial, and spectral--spatial patterns; a Vision Transformer--based tokenisation and encoding stage for modelling global contextual relationships; and a lightweight Mamba-inspired gated sequence mixing module for efficient content-aware sequence refinement without relying on quadratic-complexity self-attention. Principal Component Analysis (PCA) is employed as a preprocessing step to reduce spectral redundancy and improve computational efficiency. Extensive experiments are conducted on four benchmark hyperspectral datasets, including Houston and three UAV-borne QUH datasets (Pingan, Qingyun, and Tangdaowan). Quantitative results, evaluated using Overall Accuracy, Average Accuracy, and the Kappa coefficient, demonstrate that ConvVitMamba consistently outperforms state-of-the-art CNN-, Transformer-, and Mamba-based methods while maintaining a favourable balance between classification accuracy, model size, and inference efficiency. Ablation studies further confirm the complementary contributions of the multiscale convolutional, transformer, and Mamba-inspired components. These results indicate that the proposed framework provides an effective and efficient solution for hyperspectral image classification under both urban and natural scene settings. The source code is publicly available at \url{https://github.com/mqalkhatib/ConvVitMamba}
\end{abstract}


\section{Introduction}
Remote sensing is a multidisciplinary field that focuses on obtaining information about the Earth’s surface and atmosphere without direct contact \parencite{navalgund2007remote}. This is commonly achieved using sensors deployed on satellites \parencite{chuvieco2020fundamentals}, airborne platforms \parencite{vane1984airborne}, and unmanned aerial vehicles \parencite{yao2019unmanned}. By measuring reflected or emitted electromagnetic radiation across different regions of the spectrum, remote sensing supports consistent, wide-area, and economically efficient observation of both natural landscapes and human activities. It has become an essential tool in numerous applications such as land cover and land use mapping \parencite{anderson1976land}, environmental monitoring \parencite{li2020review}, agricultural assessment and crop monitoring \parencite{pinter2003remote}, dust cloud detection \parencite{alkhatib2012automated}, urban analysis and planning \parencite{wellmann2020remote}, and climate change studies \parencite{yang2013role}. Within this broad domain, hyperspectral imaging \parencite{bioucas2013hyperspectral} has attracted increasing interest because it captures hundreds of narrow and contiguous spectral bands for each pixel, resulting in detailed spectral signatures. Such rich spectral information enables accurate discrimination of materials and surface characteristics, making hyperspectral imagery particularly suitable for fine-level scene interpretation and high-precision classification, and positioning it as a key data source in contemporary remote sensing research.

Hyperspectral image classification is the task of assigning a semantic label to each pixel in an image, with the aim of improving data interpretation by exploiting the spectral signature of individual pixels to infer material properties. While traditional machine learning approaches such as support vector machines (SVM) \parencite{melgani2002support}, Decision Trees \parencite{goel2003classification} and Random Forest classifiers \parencite{joelsson2005random} have been widely adopted, they often struggle to model the complex and highly nonlinear relationships inherent in hyperspectral data. The strong dependencies across numerous spectral bands pose significant challenges for SVMs, especially when nonlinear kernels are required, leading to increased computational cost. Decision tree classifiers suffer from sensitivity to noise and a high risk of overfitting in the high dimensional hyperspectral feature space, which limits their generalization ability. Random Forest methods, although more robust, still struggle to capture subtle spectral interactions due to the independent nature of their tree ensembles. Moreover, these approaches rely mainly on spectral information and ignore spatial context, which ultimately constrains their classification performance.

Recent advances in deep learning, particularly with Convolutional Neural Networks (CNNs), have substantially improved hyperspectral image classification by enabling automatic learning of hierarchical feature representations \parencite{lee2017going, yu2017convolutional}. Through layered architectures, CNNs extract informative patterns directly from raw data, where shallow layers capture basic spatial structures and deeper layers learn more abstract and discriminative features, making them well suited for jointly \highlighting{modelling} spectral and spatial information \parencite{makantasis2015deep, yu2020simplified, gao2018hyperspectral, vaddi2020hyperspectral}. Despite their effectiveness, CNN based approaches still exhibit limitations in hyperspectral scenarios. Two dimensional CNNs are efficient in capturing spatial information but are less effective in \highlighting{modelling} complex spectral dependencies, while three dimensional CNNs offer richer spectral spatial representations at the cost of high computational complexity and substantial training data requirements. Moreover, deep stacks of three dimensional convolutions can complicate \highlighting{optimisation}, particularly given the limited availability of labeled hyperspectral samples \parencite{yang2020synergistic}.

Multiscale CNN based methods have played a key role in advancing hyperspectral image classification by enabling effective joint \highlighting{modelling} of spectral and spatial information. Early studies introduced multiscale 3D CNNs to capture spectral signatures and spatial structures in an end to end manner, showing clear improvements over single scale designs without relying on handcrafted features \parencite{he2017multi}. Subsequent works focused on adaptive multiscale spatial convolutions, demonstrating that extracting features at multiple spatial extents is more suitable for complex HSI scenes than fixed scale kernels \parencite{tian2018hyperspectral}. To address limited training samples, multiscale CNNs were combined with enhanced feature discrimination strategies, improving robustness and class separability \parencite{gong2019cnn}. More recent approaches emphasized efficient multiscale residual architectures and unified multiscale frameworks with spatial and channel attention, achieving strong performance with reduced complexity and improved generalisation \parencite{gao2020multiscale,wang2022unified}. Multibranch multiscale 3D CNN models further confirmed that fusing features from different scales leads to more discriminative representations for hyperspectral image classification \parencite{alkhatib2023tri}.

Vision Transformer (ViT) based models have recently emerged as an effective paradigm for hyperspectral image classification by addressing the limitations of convolutional networks in \highlighting{modelling} long range spectral dependencies. Early transformer oriented studies treated hyperspectral signatures as sequences, introducing spectral focused transformer backbones that learn local spectral relationships across neighboring bands and improve feature propagation through skip connections \parencite{hong2021spectralformer}. Along similar lines, self attention based transformer networks were proposed to jointly model spectral and spatial information using positional encoding and residual designs, demonstrating that transformer architectures can achieve competitive performance on standard HSI benchmarks \parencite{qing2021improved}. Hybrid designs soon followed, combining convolutional feature extraction with transformer encoders to capture shallow spectral spatial features while benefiting from high level semantic \highlighting{modelling}, as seen in spectral spatial tokenisation transformers and convolution enhanced hyperspectral transformers \parencite{sun2022spectral,yang2022hyperspectral}. More recent efforts focused on improving data efficiency and representation quality through self supervised pre training and hyperspectral specific attention mechanisms, showing strong gains under limited labeled data scenarios \parencite{scheibenreif2023masked}. To reduce model complexity and better balance local and global feature learning, lightweight and hybrid ViT frameworks integrating grouped convolutions, separable attention, and convolution transformer fusion strategies were introduced, achieving strong performance with fewer parameters \parencite{zhao2024hyperspectral,zhang2022convolution}. Finally, transformer models adapted from related remote sensing domains further demonstrated that combining CNN based feature extractors with localized attention mechanisms leads to robust and accurate hyperspectral image classification \parencite{alkhatib2024hsiformer}.

Despite the substantial progress achieved by CNN-based and transformer-based approaches, several challenges remain in hyperspectral image classification. CNN-driven frameworks often rely on increasingly complex multiscale designs to compensate for their limited ability to \highlighting{model} long-range spectral dependencies, which leads to deep architectures with high computational cost. Vision Transformer based models alleviate this limitation by capturing global spectral and spatial relationships through self-attention; however, they typically involve a large number of parameters and require extensive labeled data for effective training. This restricts their applicability in practical hyperspectral scenarios, where annotated samples are usually scarce. Hybrid CNN--Transformer architectures attempt to balance local inductive bias and global context \highlighting{modelling}, yet many of these designs still depend on stacked self-attention layers, resulting in high memory consumption and reduced computational efficiency.

In this context, Mamba networks have emerged as an efficient alternative for sequence \highlighting{modelling} by overcoming both the limited receptive field of CNNs and the quadratic complexity of self-attention in Transformers. Early state space models demonstrated strong theoretical capacity to capture long-range dependencies but were hindered by prohibitive computational cost, which was later addressed through structured and selective state space formulations that enable linear time complexity while preserving expressive power \parencite{gu2021efficiently,gu2024mamba}. By introducing input-dependent state transitions, Mamba enables content-aware sequence \highlighting{modelling} and achieves strong performance across multiple modalities with significantly improved efficiency compared to Transformer architectures. These advantages have recently been extended to vision tasks, where bidirectional Mamba-based backbones have shown that global visual context can be effectively \highlighting{modelled} without attention, leading to substantial gains in speed and memory efficiency \parencite{zhu2024vision}. Motivated by these properties, Mamba-based models have begun to gain attention in hyperspectral image classification, where long spectral sequences and limited labeled data pose critical challenges. Recent studies adapt Mamba to hyperspectral data through spectral or spectral--spatial tokenisation and efficient state space processing, demonstrating effective long-range spectral \highlighting{modelling} at low computational cost \parencite{yao2024spectralmamba,huang2024spectral}. Further hybrid designs integrate Mamba with convolutional feature extractors to jointly capture local spatial patterns and global spectral relationships, achieving improved performance and efficiency over conventional CNN- and Transformer-based approaches \parencite{zhang2025convmamba}.

Hyperspectral imagery presents several challenges that distinguish it from conventional image recognition tasks, including high spectral dimensionality, strong spectral redundancy, and the need to effectively model complex spectral--spatial interactions while maintaining computational efficiency. Hyperspectral imaging systems capture detailed spatial and spectral information simultaneously, enabling fine-grained material characterization but also introducing challenges related to acquisition variability and high-dimensional data processing \parencite{hong2026hyperspectral}. To address these difficulties, recent studies have explored scalable and data-driven learning frameworks. For example, SpectralGPT \parencite{hong2024spectralgpt} introduces a generative pretrained transformer designed specifically for spectral remote sensing data, enabling large-scale spatial--spectral representation learning through pretraining. Other approaches, such as LRR-Net \parencite{li2023lrr}, integrate low-rank representation principles with deep neural networks to improve feature representation and reduce reliance on manually tuned parameters. While these approaches demonstrate promising performance, hyperspectral image classification still requires models that can effectively capture both local spectral--spatial structures and long-range contextual dependencies under limited labeled data conditions.

To address the aforementioned limitations of existing CNN-, Transformer-, and sequence-\highlighting{modelling}-based approaches, this work proposes a unified architecture, termed \emph{ConvVitMamba}, for hyperspectral image classification. The proposed framework integrates complementary \highlighting{modelling} capabilities within a single pipeline, combining multiscale convolutional feature extraction, transformer-based global interaction \highlighting{modelling}, and efficient sequence refinement. In particular, the multiscale convolutional component is designed to extract features at spectral-only and spatial-only levels, complementing the conventional spectral--spatial feature extraction process and enabling richer characterisation of hyperspectral structures. The Vision Transformer component is employed to model long-range spectral--spatial dependencies through global token interactions, whereas the Mamba-inspired sequence mixing module performs efficient sequence refinement, enabling content-aware feature propagation with reduced computational overhead. By jointly combining these components, ConvVitMamba is tailored to the spectral complexity and fine spatial structure of hyperspectral data, achieving an effective balance between classification accuracy and computational efficiency. Unlike large-scale pretrained frameworks such as SpectralGPT, which rely on extensive pretraining and large datasets, or specialised architectures such as LRR-Net that emphasise low-rank representation learning, the proposed ConvVitMamba focuses on lightweight yet effective spectral--spatial \highlighting{modelling} by integrating convolutional, transformer, and Mamba-based sequence mechanisms within a unified and computationally efficient architecture.

The primary contributions of this article are summarised as follows:
\begin{enumerate}
    \item A unified hyperspectral image classification architecture, termed \emph{ConvVitMamba}, is proposed to jointly exploit convolutional feature extraction, transformer-based global interaction \highlighting{modelling}, and Mamba-inspired sequence refinement within a single spectral--spatial learning framework.
    
    \item A hyperspectral-aware multiscale convolutional feature extractor is designed using parallel 3D convolutional branches to capture complementary spectral-only, spatial-only, and joint spectral--spatial representations at different receptive fields, improving feature diversity while preserving computational efficiency.
    
    \item A lightweight Mamba-inspired gated sequence mixing module is introduced to refine token representations after transformer interaction, enabling efficient long-range dependency \highlighting{modelling} without relying on deep stacks of self-attention layers.
    
    \item Extensive experiments conducted on multiple benchmark hyperspectral datasets demonstrate that the proposed architecture achieves a favourable balance between classification accuracy and computational efficiency, outperforming representative CNN-, Transformer-, and Mamba-based approaches while maintaining a compact model size.
\end{enumerate}

The structure of this paper is as follows:  Section \ref{sec:Framework} details the proposed ConvVitMamba model. Section \ref{sec:results} presents the experimental setup and results. Finally, Section \ref{sec:conclusion} discusses the conclusions and outlines directions for future research.

\section{Methodology} 
\label{sec:Framework}

\subsection{Data Preprocessing}
Dimensionality reduction plays a crucial role in enabling efficient processing of hyperspectral images within deep learning frameworks. In this work, Principal Component Analysis (PCA) \parencite{ali2019analysis} is adopted as a preprocessing step to address the high spectral dimensionality of HSI data. PCA is a well established unsupervised technique that reduces data dimensionality while preserving the most informative variations in the original feature space. It exploits the strong correlation that typically exists among neighboring spectral bands, which often carry redundant information about the same materials. By applying an orthogonal transformation, PCA projects the original correlated spectral variables onto a new set of linearly uncorrelated components, referred to as principal components. For an HSI cube of size $H \times W \times C$, where $H$ and $W$ denote spatial dimensions and $C$ represents the number of spectral bands, PCA compresses the spectral dimension to $P$, producing a transformed cube of size $H \times W \times P$, with $P \ll C$.

\subsection{Architecture of the ConvViTMamba Model}
\begin{figure} [t!]
   \centering
   \includegraphics[clip=true, trim = 20 10 20 5,width= .85\linewidth]{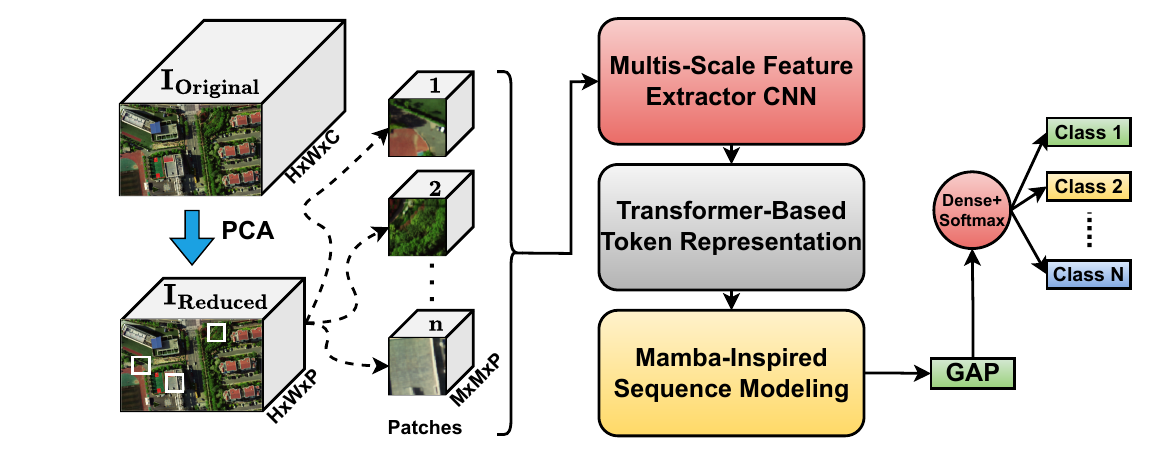}   
   \vspace{-1em}
    \caption{ \label{fig:model} Architecture of the proposed model.}
\end{figure}

This article introduces a novel model for hyperspectral image (HSI) classification, whose overall architecture is illustrated in Figure~\ref{fig:model}. The proposed \textit{ConvViTMamba} framework integrates a multi-scale 3D convolutional feature extractor, a Vision Transformer for token-level representation learning, and a lightweight Mamba-inspired gated sequence mixing module, followed by fully connected layers for classification. The convolutional, transformer, and sequence mixing components collaboratively serve as hierarchical feature extractors, capturing spectral--spatial characteristics and inter-token dependencies, while the final prediction is produced by the fully connected layers.

The feature extraction process starts by capturing the spectral–spatial characteristics of the processed HSI data through three parallel streams of 3D convolutional blocks, each composed of two consecutive 3D convolutional layers with different kernel configurations to model spectral, spatial, and joint spectral–spatial information at multiple scales. The resulting feature maps are fused and reshaped into a sequence of patch tokens, which are then processed by a Vision Transformer to capture global contextual relationships and long-range dependencies among tokens. To further enhance the token representations, a lightweight Mamba-inspired gated sequence mixing module is applied, utilizing input-dependent gating and convolution-based token mixing to efficiently refine inter-token dependencies, particularly for the short token sequences typical in patch-based HSI classification. Finally, the refined features are aggregated and fed into fully connected layers to generate the final classification results, with further architectural details provided in the subsequent sections.

\subsubsection{Multis-Scale Feature Extractor CNN}
Currently, most hyperspectral image (HSI) classification methods rely on either 2D-CNN or 3D-CNN architectures. While 2D-CNNs are effective in \highlighting{modelling} spatial structures, they are limited in their ability to fully exploit the rich spectral information inherent in hyperspectral data. On the other hand, 3D-CNNs jointly model spatial and spectral information; however, treating both dimensions simultaneously may lead to suboptimal feature representation when the characteristics of each dimension are not explicitly distinguished. To address these limitations, a three-branch feature fusion network is introduced, comprising a spatial feature extractor, a spectral--spatial feature extractor, and a spectral-only feature extractor. This design aims to enhance feature discrimination by explicitly \highlighting{modelling} complementary spatial and spectral characteristics and improving the overall representation capability.

\begin{figure} [t!]
   \centering
   \includegraphics[clip=true, trim = 20 5 20 5,width= .7\linewidth]{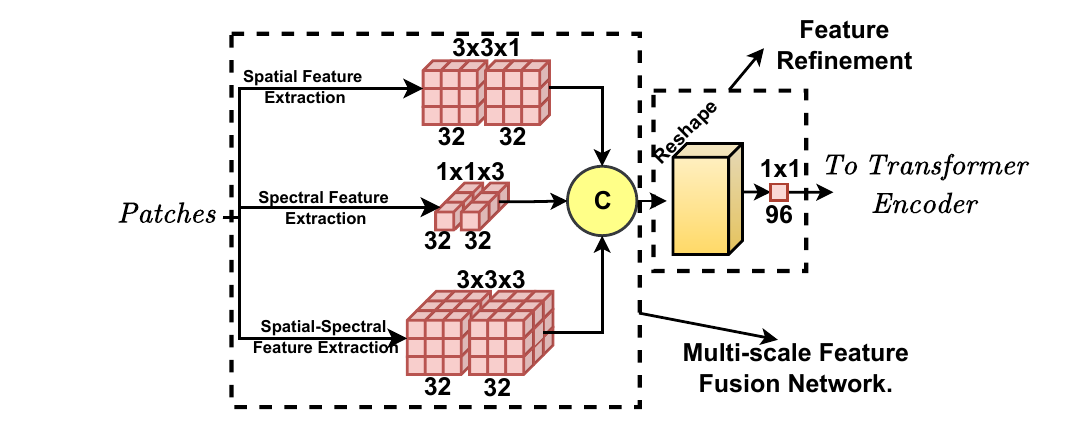}   
   \vspace{-1em}
    \caption{ \label{Fig:MS_FE} Multi-scale Feature Extractor Network.}
\end{figure}

In the proposed framework, spatial features are extracted using $3 \times 3 \times 1$ 3D convolution kernels, which focus on local spatial patterns within individual spectral channels while preserving spectral integrity as shown in Figure \ref{Fig:MS_FE}. The use of compact kernels contributes to improved computational efficiency without compromising feature quality \parencite{pei2022small}. Spectral features are captured using $1 \times 1 \times 3$ 3D convolution kernels, which model inter-channel relationships by operating on groups of three adjacent spectral bands at the pixel level. In addition, joint spectral--spatial features are extracted using $3 \times 3 \times 3$ 3D convolution kernels, which simultaneously encode local spatial context and spectral interactions across neighboring bands. This strategy has been shown to outperform approaches that rely solely on spatial or spectral feature extraction \parencite{hamida20183}. Each kernel configuration serves a distinct purpose: spatial kernels operate on single channels, spectral kernels capture dependencies among adjacent bands, and spectral--spatial kernels integrate both dimensions. Importantly, feature extraction is performed on subsets of channels rather than across the entire spectral dimension at once, allowing each branch to focus on its designated role while collectively contributing to a comprehensive feature representation.

Although multiscale feature fusion architectures are widely adopted in HSI classification, the selection of appropriate kernel sizes remains an open research question. For example, the method proposed in \parencite{Roy2020} employs cascaded convolution kernels of sizes $(3,3,7)$, $(3,3,5)$, and $(3,3,3)$, with filter dimensions determined empirically through extensive experimentation. In the proposed model, kernel sizes are carefully selected to effectively capture discriminative features within each branch while maintaining architectural simplicity. Furthermore, the number of filters in each convolutional layer is fixed at 32 to avoid excessive model complexity and ensure computational efficiency, as experimental evaluations indicate that increasing the number of filters beyond this value does not yield noticeable performance gains. The resulting multiscale feature maps are then concatenated and reshaped into a unified representation, followed by a $1\times1$ convolutional filter that performs channel-wise feature fusion and further refines the extracted features before \highlighting{tokenisation}.

\subsubsection{Transformer-Based Token Representation}
Hyperspectral images (HSIs) contain rich spectral signatures alongside detailed spatial information, making them highly informative for land-cover analysis and scene understanding. Convolutional neural networks (CNNs) have been widely adopted for HSI classification due to their strong capability in extracting local spatial patterns. However, conventional 2D-CNNs primarily focus on spatial structures and often fail to fully exploit the high-dimensional spectral information. Although 3D-CNNs process spectral and spatial dimensions jointly, treating both dimensions in a uniform manner may limit their ability to explicitly model long-range spectral dependencies and global contextual relationships.

To address these challenges, transformer-based models have recently been introduced into HSI classification. Owing to their self-attention mechanism, transformers are well suited for capturing long-range dependencies and global contextual information across the image. Vision Transformers (ViTs) adapt this concept to visual data by dividing an image into a set of non-overlapping patches, which are then processed as a sequence of tokens. This representation enables the model to jointly reason about spatial regions and spectral responses over the entire scene. Furthermore, hybrid architectures that combine convolutional feature extractors with transformers have demonstrated improved performance by benefiting from both local feature \highlighting{modelling} and global context learning.

In a standard ViT framework for hyperspectral data, the input image is partitioned into patches of size $P \times P \times C$, where $P$ denotes the spatial patch size and $C$ is the number of spectral bands. Each patch is flattened and projected into a latent feature space through a linear embedding, and positional encodings are added to preserve spatial ordering. The resulting sequence of embedded patches serves as the input to the transformer encoder, where layer normalization and self-attention are applied to model relationships among all tokens in the sequence.

\begin{figure} [t!]
   \centering
   \includegraphics[clip=true, trim = 20 5 20 5,width= .95\linewidth]{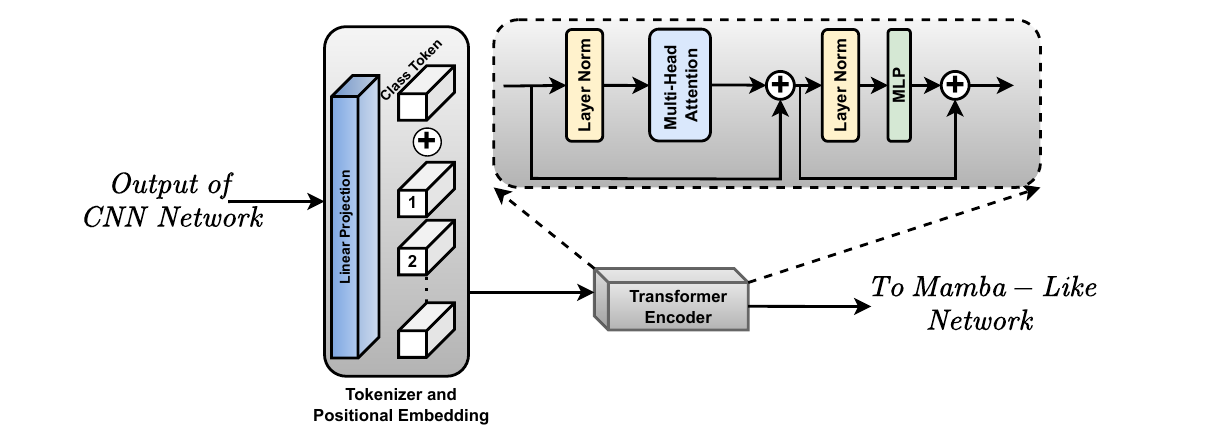}   
   \vspace{-1em}
    \caption{ \label{Fig:ViT} Transformer-Based Token Representation.}
\end{figure}

The self-attention mechanism computes interactions between patch tokens by projecting them into query, key, and value representations and estimating their pairwise relevance. This operation enables the model to selectively emphasize informative spectral–spatial regions while suppressing less relevant ones. To enhance representational capacity, multi-head attention is employed, allowing the model to attend to different aspects of the data in parallel. Following the attention stage, a feed-forward multi-layer perceptron further refines the token representations by introducing nonlinearity and channel-wise feature interactions. Figure \ref{Fig:ViT} shows the block diagram associated with the Transformer-Based Token Representation

Finally, the refined patch features produced by the transformer encoder are not directly used for classification. Instead, the output token sequence is further processed by a lightweight Mamba-inspired sequence \highlighting{modelling} block, which aims to enhance inter-token interactions through gated convolutional mixing. This additional stage refines the transformer features before aggregation and classification, and is particularly effective for \highlighting{modelling} the short token sequences typical in patch-based hyperspectral image analysis. The detailed formulation and design of this Mamba-style block are presented in the following section.

\subsubsection{Mamba-Inspired Sequence Modelling}
\begin{figure}[t!]
   \centering
   \includegraphics[clip=true, trim=20 5 20 5, width=0.95\linewidth]{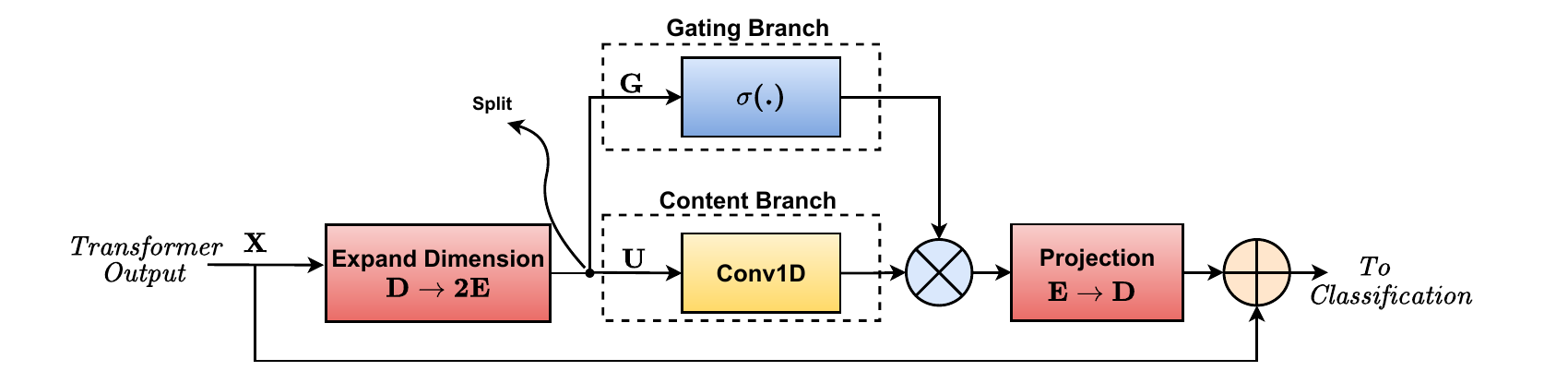}
   \vspace{-1em}
   \caption{Block diagram of the proposed Mamba-inspired sequence \highlighting{modelling} block.}
   \label{Fig:Mamba}
\end{figure}

While the transformer encoder is effective in capturing global contextual relationships among patch tokens, its output can be further refined to strengthen inter-token interactions and improve feature consistency. To this end, a lightweight Mamba-inspired sequence \highlighting{modelling} block is introduced after the transformer stage. This module, presented in Figure~\ref{Fig:Mamba}, is designed to enhance token representations through gated sequence mixing, while maintaining low computational complexity, making it particularly suitable for the short token sequences encountered in patch-based hyperspectral image classification. It should be noted that the proposed block is designed as a lightweight adaptation inspired by Mamba-style sequence \highlighting{modelling} rather than a direct implementation of the full selective state-space formulation. The goal is to retain the key idea of efficient, input-adaptive sequence interaction while maintaining a compact structure suitable for hyperspectral learning scenarios.

Let $X \in \mathbb{R}^{B \times T \times D}$ denote the output of the transformer encoder, where $B$ is the batch size, $T$ is the number of tokens, and $D$ is the embedding dimension. The sequence \highlighting{modelling} process begins with a linear projection that expands the feature dimension and splits the representation into two parallel branches, namely a content branch and a gating branch:
\begin{equation}
[U, G] = X W,
\end{equation}
where $W \in \mathbb{R}^{D \times 2E}$ is the projection matrix and $E = \alpha D$ denotes the expanded feature dimension controlled by the expansion factor $\alpha$.
The content branch $U$ is processed by a one-dimensional convolution along the token dimension to enable local sequence mixing:
\begin{equation}
U_c = \mathrm{Conv1D}(U),
\end{equation}
followed by a GELU activation function to introduce nonlinearity. This convolutional mixing operation provides efficient token interaction along the sequence dimension, enabling contextual information propagation while preserving linear computational complexity with respect to the token length. In parallel, the gating branch $G$ is passed through a sigmoid function to generate adaptive gating coefficients. The two branches are then combined through element-wise multiplication:
\begin{equation}
U_m = \mathrm{GELU}(U_c) \odot \sigma(G),
\end{equation}
allowing the model to dynamically control the flow of information across tokens.

The gated features are subsequently projected back to the original embedding dimension $D$ using a linear transformation, and a residual connection is applied to preserve the original transformer features:
\begin{equation}
Y = X + U_m W_o,
\end{equation}
where $W_o \in \mathbb{R}^{E \times D}$ is the output projection matrix. This residual formulation facilitates stable \highlighting{optimisation} and ensures that the sequence \highlighting{modelling} block refines, rather than overwrites, the transformer representations.

Overall, the proposed sequence \highlighting{modelling} block acts as an efficient refinement module that complements the transformer encoder by \highlighting{modelling} short-range token dependencies through gated convolutional operations. The design is inspired by the core principles of Mamba and selective state space models, particularly the use of content-dependent gating, linear-time sequence processing, and residual refinement to enhance representation quality without relying on self-attention. While the present formulation adopts a simplified convolutional realization rather than an explicit state space parameterization, it preserves the key Mamba philosophy of efficient, input-adaptive sequence mixing, making it well aligned with the requirements of hyperspectral image classification. This design choice allows the proposed framework to benefit from efficient sequence \highlighting{modelling} while avoiding the computational overhead associated with full state-space parameterization, which is less critical for the relatively short token sequences typically encountered in patch-based hyperspectral classification.
\section{Experiments and Analysis}
\label{sec:results}
This section employs four widely used hyperspectral image datasets, namely QUH-Pingan, QUH-Qingyun, and QUH-Tangdaowan~\parencite{fu2023three}, together with the Houston dataset~\parencite{debes2014hyperspectral}, to assess the effectiveness of the proposed method under diverse experimental settings. A comprehensive evaluation is conducted by presenting both qualitative classification maps and quantitative performance metrics, complemented by comparisons with state-of-the-art approaches.

\subsection{Datasets}

The proposed method was evaluated on four widely used hyperspectral datasets, namely (a) Houston, (b) QUH Pingan, (c) QUH Qingyun, and (d) QUH Tangdaowan. These datasets cover diverse land-cover scenarios and acquisition conditions, providing a comprehensive benchmark for assessing the robustness and generalization capability of the proposed approach. 

\begin{figure} [t!]
   \centering
   \includegraphics[clip=true, trim = 20 10 20 5,width= .5\linewidth]{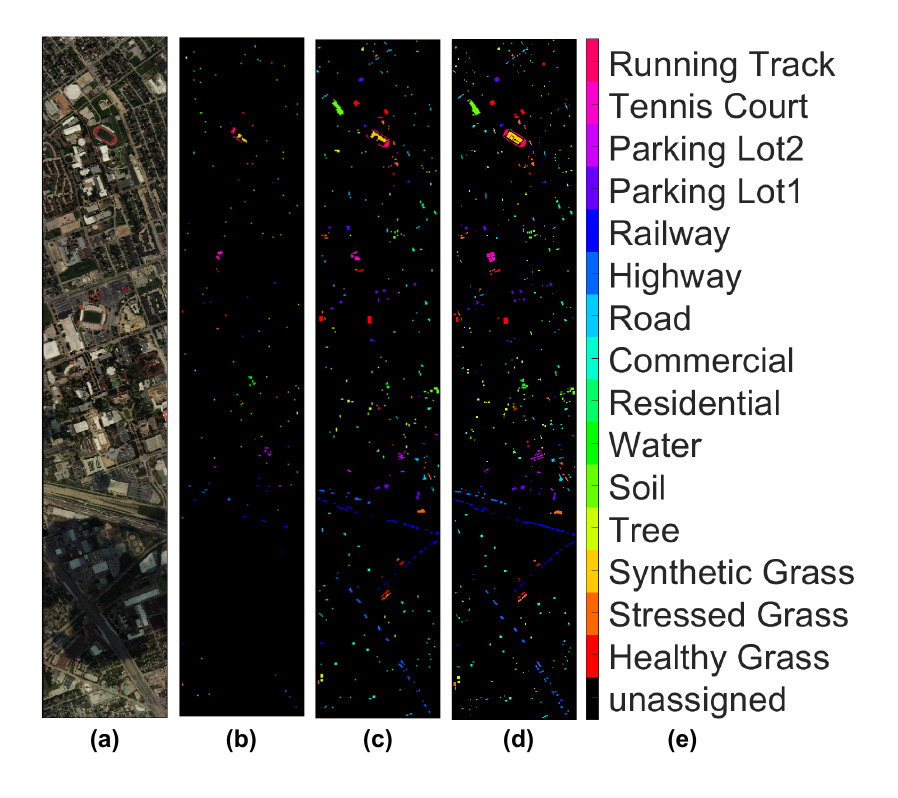}   
   \vspace{-1em}
    \caption{ \label{fig:Datesets_H} Houston Dataset. (a) RGB Image; (b) Train Image; (c) Test Image; (d) Full Reference Map and (e) Class Labels.}
\end{figure}

\begin{figure} [t!]
   \centering
   \includegraphics[clip=true, trim = 20 10 20 5,width= .99\linewidth]{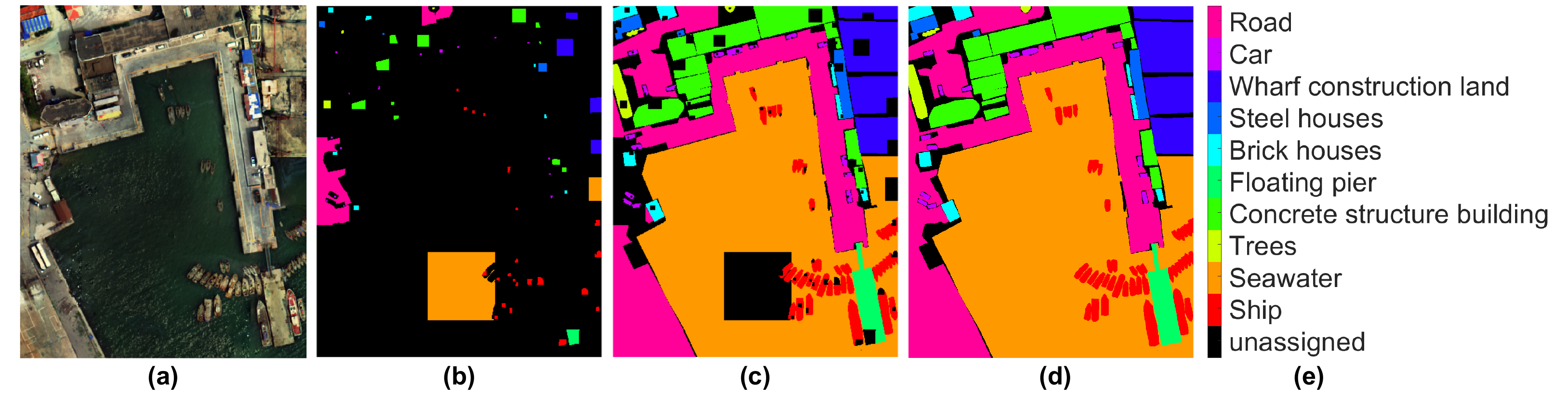} 
   \vspace{-1em}
    \caption{ \label{fig:Datesets_P} QUH Pingan Dataset. (a) RGB Image; (b) Train Image; (c) Test Image; (d) Full Reference Map and (e) Class Labels.}
\end{figure}

\begin{figure} [t!]
   \centering
   \includegraphics[clip=true, trim = 20 10 20 5,width= .99\linewidth]{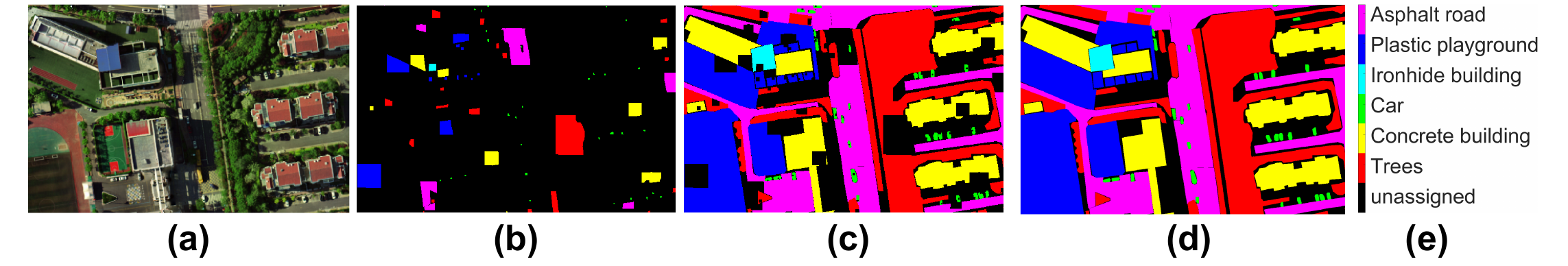}   
   \vspace{-1em}
    \caption{ \label{fig:Datesets_Q} QUH Qingyun Dataset. (a) RGB Image; (b) Train Image; (c) Test Image; (d) Full Reference Map and (e) Class Labels.}
\end{figure}

\begin{figure} [t!]
   \centering
   \includegraphics[clip=true, trim = 20 10 20 5,width= .99\linewidth]{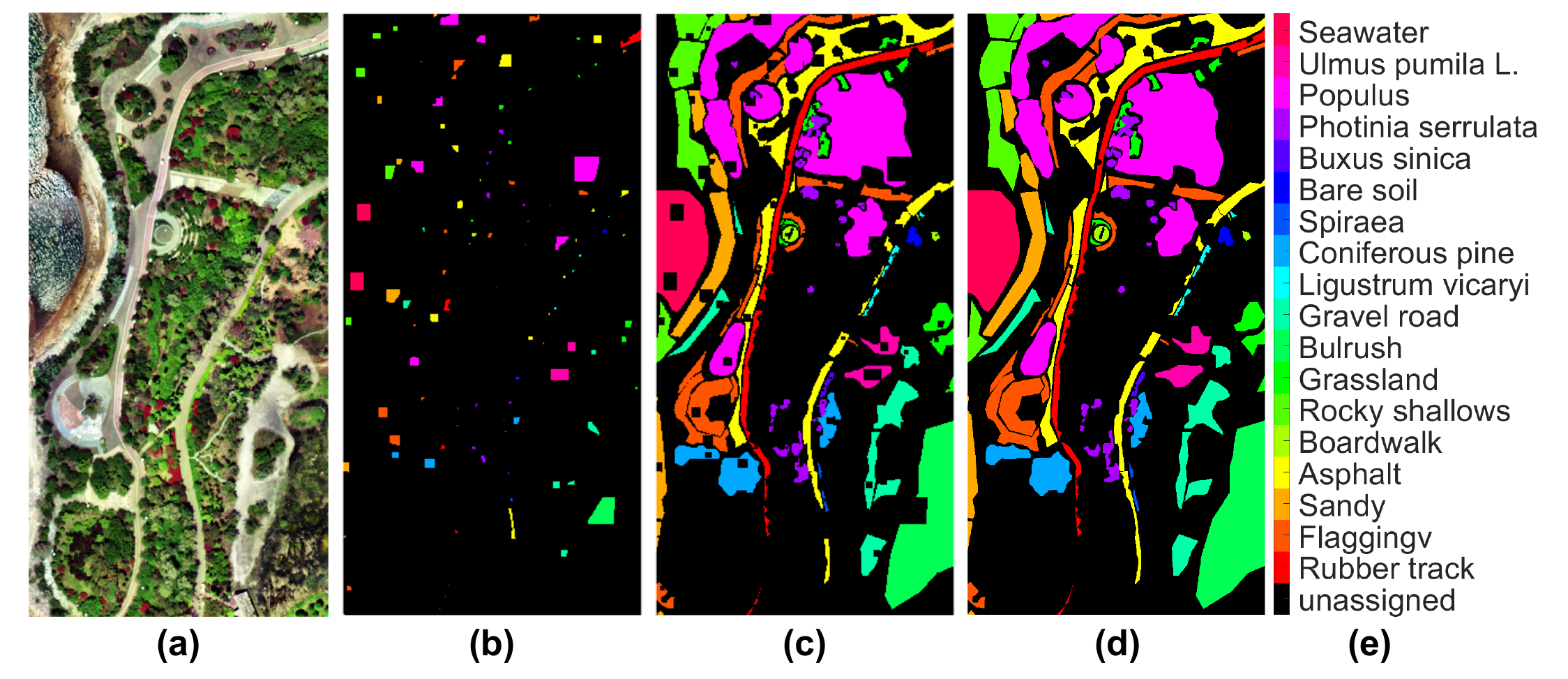}   
   \vspace{-1em}
    \caption{ \label{fig:Datesets_T} QUH Tangdaowan Dataset. (a) RGB Image; (b) Train Image; (c) Test Image; (d) Full Reference Map and (e) Class Labels.}
\end{figure}

\begin{enumerate}

\item[(a)] \textbf{Houston:}  
The Houston dataset was acquired on 23 June 2012 over the University of Houston campus and its surrounding area by the NSF-funded center for Airborne Laser Mapping (NCALM). The hyperspectral data were collected with an average sensor height of approximately 1600~m above ground level, resulting in a spatial resolution of 2.5~m. The dataset consists of 144 spectral bands covering the wavelength range from 380 to 1050~nm, with an image size of $1905 \times 349$ pixels. The RGB composite image, together with the training/testing split and the complete reference class map for the Houston dataset, are shown in Figure~\ref{fig:Datesets_H}.
\newline

\item[(b)] \textbf{QUH Pingan:}  
The QUH Pingan dataset was collected on 19 May 2021 at the Huangdao Pingan Passenger Ship Terminal in Qingdao, China. The hyperspectral data were acquired using a UAV platform operating at a height of 200~m above ground level, resulting in a spatial resolution of approximately 0.10~m. The dataset consists of 176 spectral bands covering the wavelength range from 400 to 1000~nm, with an image size of $1230 \times 1000$ pixels. The RGB composite image, together with the training/testing split and the complete reference class map for the QUH Pingan dataset are shown in Figure~\ref{fig:Datesets_P}. 
\newline

\item[(c)] \textbf{QUH Qingyun:}  
The QUH Qingyun dataset was surveyed on 18 May 2021 in the vicinity of the Qingyun Road primary school and the surrounding residential area in Qingdao, China. The hyperspectral images were captured using a UAV platform operating at a height of 300~m above ground level, resulting in a spatial resolution of approximately 0.15~m. The dataset comprises 270 spectral bands spanning the wavelength range from 400 to 1000~nm, with an image size of $880 \times 1360$ pixels. The RGB composite image, together with the training/testing split and the complete reference class map for the QUH Qingyun dataset are presented in Figure~\ref{fig:Datesets_Q}. 
\newline

\item[(d)] \textbf{QUH Tangdaowan:}  
The QUH Tangdaowan dataset was surveyed on 18 May 2021 in Tangdao Bay National Wetland Park, Qingdao, China. The hyperspectral data were acquired using a UAV platform operating at a height of 300~m above ground level, resulting in a spatial resolution of approximately 0.15~m. The dataset consists of 176 spectral bands covering the wavelength range from 400 to 1000~nm, with an image size of $1740 \times 860$ pixels. The RGB composite image, together with the training/testing split and the complete reference class map for the QUH Tangdaowan dataset are illustrated in Figure~\ref{fig:Datesets_T}.
\end{enumerate}

For the QUH datasets and the Houston dataset, training samples were selected following the protocols described in~\parencite{fu2023three} and~\parencite{debes2014hyperspectral}, respectively. This sampling strategy minimizes spatial overlap between training and testing samples, thereby reducing information leakage. A validation set was further constructed by randomly selecting 30\% of the training samples for model selection and hyperparameter tuning. 
Adhering to these established protocols ensures consistency with widely adopted experimental settings in hyperspectral image classification and enables fair comparison with previously reported methods. Exploring additional scenarios such as cross-dataset generalization or extremely low training sample ratios remains an interesting direction for future work.

Although the QUH-Pingan, QUH-Qingyun, and QUH-Tangdaowan datasets originate from the same geographic region (West Coast New Area, Qingdao City, Shandong Province, China)~\parencite{fu2023three}, they were acquired under different imaging conditions, including variations in sensor type and ground sampling distance caused by different acquisition altitudes. As a result, these datasets exhibit different spatial resolutions and scene characteristics. In contrast, the Houston dataset represents an urban hyperspectral scene acquired in Houston, USA, providing an additional evaluation scenario with substantially different environmental and spectral properties.

\subsection{Experimental Configuration}
All experiments were conducted using Python~3.9 and TensorFlow~2.10 on a Windows~11 workstation equipped with 64~GB of RAM and an NVIDIA GeForce RTX~2080 GPU with 8~GB of VRAM. The Adam optimizer was employed with an initial learning rate of \(1 \times 10^{-3}\), and a batch size of 32 was used. Each model was trained for a maximum of 100 epochs. 

During training, a validation-driven \highlighting{optimisation} and model selection strategy was applied. The network weights corresponding to the highest validation accuracy were automatically saved using a checkpointing mechanism, ensuring that the best-performing model was retained even if subsequent epochs led to overfitting. In addition, an adaptive learning rate scheduling scheme was employed, whereby the learning rate was reduced by a factor of 0.5 whenever the validation accuracy did not improve for ten consecutive epochs, with a minimum learning rate of \(1 \times 10^{-5}\). This dynamic adjustment facilitated stable convergence and improved generalization. To mitigate the influence of random weight initialization and training stochasticity, each experiment was repeated ten times, and the final results are reported as the mean and standard deviation over these independent runs.

\subsection{Evaluation Metrics}
Assessing the effectiveness of classification performance requires comparing the predicted class maps with the corresponding reference (ground-truth) data. Relying solely on visual inspection to verify pixel-wise accuracy is inherently subjective and may not provide a comprehensive evaluation. Therefore, quantitative assessment is adopted to ensure objective and reliable performance analysis. In this work, three widely used evaluation metrics are employed: Overall Accuracy (OA), Average Accuracy (AA), and the Kappa coefficient ($\kappa$).

Overall Accuracy (OA) is defined as the ratio of correctly classified pixels to the total number of samples. Average Accuracy (AA) represents the mean of the classification accuracies computed independently for each class. The Kappa coefficient ($\kappa$) measures the level of agreement between the predicted classification map and the ground truth while accounting for chance agreement, with values ranging from 0 to 1. A value of $\kappa = 1$ indicates perfect agreement, whereas $\kappa = 0$ denotes complete disagreement. In practice, a Kappa value greater than or equal to 0.80 is generally interpreted as substantial agreement, while values below 0.40 indicate poor classification performance~\parencite{cohen1960coefficient}.

\subsection{Parameter Analysis}
Hyperparameter selection plays a crucial role in achieving reliable performance and good generalization in deep learning models. In this study, the effects of patch size and the number of principal components on classification accuracy are systematically investigated across multiple hyperspectral image datasets. The patch size defines the spatial neighborhood used to extract spectral and spatial information for each pixel, where large patches may introduce spectral mixing from adjacent classes and small patches may fail to capture sufficient spatial context. To analyze this trade off, patch sizes of $\{5 \times 5, 7 \times 7, 9 \times 9, 11 \times 11, 13 \times 13, 15 \times 15, 17 \times 17\}$ were evaluated. In parallel, dimensionality reduction was performed by varying the number of retained principal components in $\{10, 15, 20, 25, 30, 35\}$, where increasing the number of components generally preserves more discriminative spectral information, while excessive components introduce redundancy and noise that can negatively impact classification accuracy.

\begin{figure} [t!]
   \centering
   \includegraphics[clip=true, trim = 20 10 20 5,width= .75\linewidth]{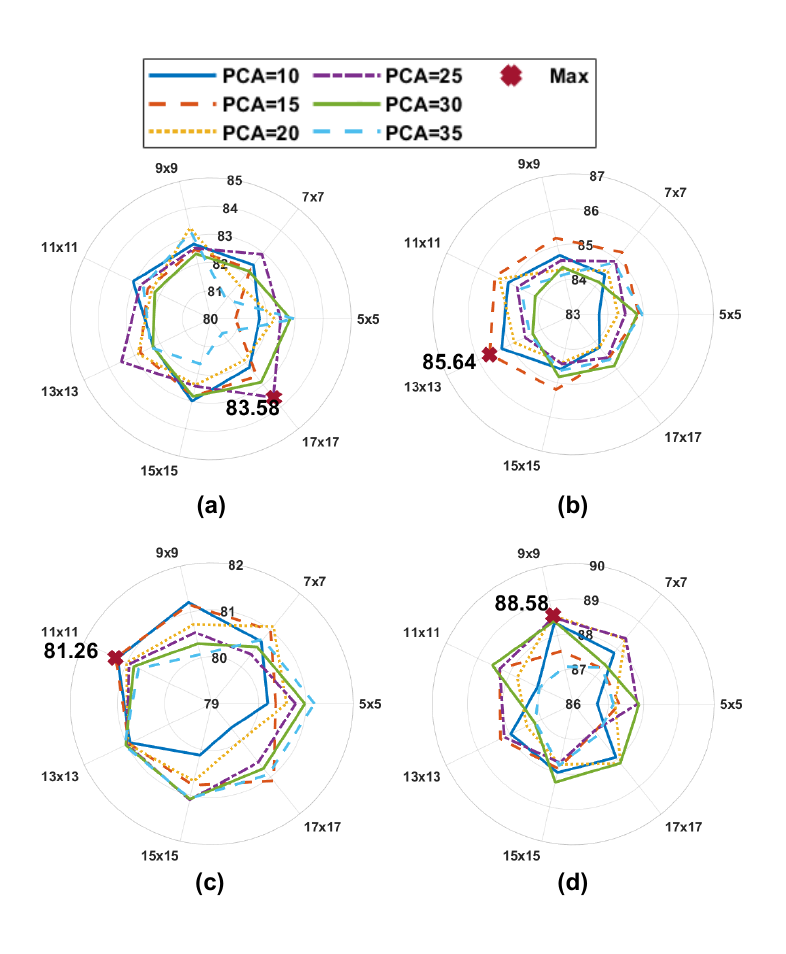}   
   \vspace{-2em}
    \caption{ \label{fig:PCA_vs_win} Parameter Radar (Spider) plots of Overall Accuracy on four datasets. (a) Houston. (b) QUH-Pingan. (c) QUH-Qingyun. (d) QUH-Tangdaowan.}
\end{figure}

Figure \ref{fig:PCA_vs_win} presents radar plots depicting the variation in classification accuracy across different combinations of patch sizes and PCA dimensionalities for the four datasets. For each dataset, a distinct patch size–PCA configuration was found to be optimal. Specifically, the best performance was achieved using a $17 \times 17$ patch size with PCA = 25 for Houston (83.52\%), $13 \times 13$ with PCA = 15 for QUH-Pingan (85.64\%), $11 \times 11$ with PCA = 15 for QUH-Qingyun (81.26\%), and $9 \times 9$ with PCA = 20 for QUH-Tangdaowan (88.58\%).

The higher performance fluctuations observed for the Tangdaowan dataset are mainly due to its severely imbalanced class distribution, where several classes contain very limited training samples. This imbalance increases the sensitivity of the model to variations in patch size and PCA dimensionality, as larger patches may introduce class mixing for minority classes, while inappropriate PCA compression can either remove discriminative information or amplify noise. Consequently, these factors jointly lead to less stable performance across different configurations. Table \ref{tab:winVsPCA} shows a summary of the Optimal Patch size along with the number of PCA count.
\input{tab_winSizeVsPCA}

\input{tab_ablation}

\subsection{Ablation Study}
Table~\ref{tab:Ablation_AllDatasets} provides a detailed ablation analysis of the three core modules in ConvViTMamba, namely the multiscale feature extractor (MS\_FE), the transformer encoder stack (ViT), and the Mamba style sequence block. Following the experimental protocol described earlier, each ablation configuration was also evaluated over ten independent runs with different random initializations. The values reported in Table~\ref{tab:Ablation_AllDatasets} correspond to the mean performance across these runs. Across the repeated runs, the observed standard deviations remained small for all ablation configurations, indicating that the performance differences between model variants are stable and not caused by random training fluctuations. Across all datasets, the full model consistently achieves the best performance, confirming that the three components are complementary rather than redundant. For Houston, the complete configuration yields an OA of 83.30\%, AA of 86.16\%, and $\kappa$ of 81.91, while removing MS\_FE produces a clear degradation in all metrics (82.65\% OA and 81.05 $\kappa$), highlighting the importance of multiscale spectral and spatial feature construction before tokenization. A larger drop is observed when the ViT stack is removed (81.90\% OA and 80.10 $\kappa$), indicating that self attention based global interaction among patches remains a primary driver for robust discrimination in this scene. Removing the Mamba block also reduces performance (82.40\% OA and 80.95 $\kappa$), but the decline is smaller than removing ViT, suggesting that sequence \highlighting{modelling} acts as a refinement stage that strengthens token level representations after the transformer and MLP head. Similar \highlighting{behaviour} is observed for QUH Pingan, where the full model achieves 85.64\% OA and 79.11 $\kappa$, and performance consistently decreases when any component is disabled. The removal of ViT leads to the largest reduction in $\kappa$ (from 79.11 to 77.60), implying that global contextual \highlighting{modelling} is critical to preserve class separability in this dataset, whereas removing MS\_FE or Mamba causes slightly smaller but still systematic drops. For QUH Qingyun, the reductions follow the same trend, with the ViT ablation producing the lowest OA and $\kappa$ among the variants, which reinforces the role of self attention in stabilizing performance when spectral characteristics alone are insufficient for discrimination. In contrast, Tangdaowan shows a comparatively stronger sensitivity to MS\_FE and ViT, where disabling MS\_FE reduces OA from 88.58\% to 87.80\% and disabling ViT reduces it further to 87.10\%, while $\kappa$ drops from 87.07 to 86.20 and 85.30, respectively. This pattern is consistent with the complex spatial layout and potential class mixing in Tangdaowan, where multiscale feature extraction mitigates local ambiguity and the ViT stack enforces global consistency across patch tokens. Notably, across all datasets, the Mamba removal causes a smaller but persistent performance loss, which suggests that the Mamba style block contributes additional sequential dependency \highlighting{modelling} and feature smoothing that improves robustness without replacing the transformer. Overall, the ablation results verify that MS\_FE strengthens local spectral and spatial representations, ViT provides essential global token interaction and yields the largest gains, and the Mamba block offers consistent refinement that further improves accuracy and agreement, as reflected by the systematic improvements in OA, AA, and $\kappa$ when it is included.

\subsection{Comparison with Other Methods}
To evaluate the performance of the proposed model, a comprehensive comparison is carried out with representative machine learning and deep learning approaches reported in recent literature. The evaluation includes classical machine learning methods such as SVM~\parencite{melgani2004classification} and MLP~\parencite{thakur2017hyper}, as well as convolutional neural network based models, including 2D-CNN~\parencite{chen2016deep}, 3D-CNN~\parencite{hamida20183}, and HybridSN~\parencite{Roy2020}. In addition, transformer based architectures that have shown strong capability in \highlighting{modelling} long range dependencies are considered, namely the Vision Transformer (ViT)~\parencite{dosovitskiy2020image}, DiffFormer~\parencite{ahmad2025diffformer}, and SimPoolFormer~\parencite{roy2025simpoolformer}. Furthermore, Kolmogorov Arnold Networks are represented by HybridKAN~\parencite{jamali2024learn}, while recent state space and Mamba based approaches are represented by MorphMamba~\parencite{ahmad2025spatial} and WaveMamba~\parencite{ahmad2024wavemamba}. This diverse set of baselines enables a thorough assessment of the proposed method against CNNs, transformer models, Kolmogorov Arnold Networks, and Mamba Networks. All comparison methods were implemented using the optimized hyperparameters reported in their original works to ensure a fair and representative evaluation.

\subsubsection{Model Size and Computational Efficiency}
The model size, arithmetic cost, and end to end inference time over the full Houston image are summarized in Table~\ref{tab:complexity}, including the number of parameters, floating point operations (FLOPs), multiply and accumulate operations (MACs), and the time required to predict all pixels in the scene. In general, smaller baselines such as MLP, 2D-CNN, 3D-CNN, and HybridSN exhibit short inference times between 0:36 and 0:44, which is consistent with their relatively low FLOPs and MACs. In contrast, transformer based models tend to increase runtime due to heavier token processing, where ViT reaches 4:18 and SimPoolFormer reaches 4:55, consistent with their higher computational requirements, particularly for SimPoolFormer with 57,497,159 FLOPs and 28,423,767 MACs. The proposed model requires 384,010 parameters with 22,706,496 FLOPs and 11,351,808 MACs, while achieving an inference time of 2:16, indicating that it remains practically efficient even when full image prediction is performed.

It is also observed that Mamba based networks can be lightweight in parameter count, yet still incur long inference time when deployed in a full scene setting. For example, MorphMamba uses only 67,650 parameters with moderate arithmetic cost of 8,592,682 FLOPs and 4,294,396 MACs, but still requires 6:04 for full image inference in Table~\ref{tab:complexity}. This \highlighting{behaviour} is consistent with the design choice of scan based token processing used in Mamba style models, where patch cross scanning and bidirectional spectral scanning are used to capture spatial and spectral context, and the computational time is strongly influenced by patch size and related scanning settings. WaveMamba shows an even more pronounced effect, despite having only 61,711 parameters with 4,361,728 FLOPs and 2,178,944 MACs, its inference time reaches 22:00 in Table~\ref{tab:complexity}. The WaveMamba design applies a Haar wavelet transform that decomposes features into multiple subchannels, followed by a state transition that is computed sequentially across the token sequence, which increases per patch overhead during full image prediction.  
\input{tab_complexity}

Although certain lightweight baselines achieve lower parameter counts, the proposed model maintains competitive computational complexity while achieving higher classification accuracy, as demonstrated in the subsequent experimental results. This highlights a favorable trade-off between model efficiency and predictive performance.

\subsubsection{Results of Houston Dataset}
Table~\ref{tab:Houston_Results} summarizes the classification performance across individual land cover classes for the Houston dataset, providing a detailed view of class wise \highlighting{behaviour} in addition to the overall accuracy. Clear performance variations are observed across classes with different spectral and spatial characteristics. Classes such as Soil, Tennis Court, and Synthetic Grass are consistently classified with high accuracy by most deep learning models. For instance, Soil achieves accuracies of 99.96\% with 3D CNN, 99.82\% with HybridSN, 98.55\% with SimPoolFormer, and 99.97\% with the proposed ConvVitMamba. Similarly, Tennis Court reaches 99.92\% using 3D CNN, 99.43\% with HybridSN, 99.52\% with SimPoolFormer, and 99.91\% with the proposed method. Synthetic Grass further highlights the advantage of deep spectral spatial \highlighting{modelling}, where 3D CNN attains 96.12\% and ConvVitMamba achieves the highest accuracy of 99.42\%, compared to only 58.99\% with MLP. In contrast, spectrally similar and structurally complex classes such as Commercial, Highway, Railway, and Running Track remain challenging. For Commercial areas, SVM achieves only 36.79\%, while the proposed model improves performance to 73.34\%. Highway remains difficult across most methods, with accuracies generally below 50\%, although SimPoolFormer and ConvVitMamba reach 56.72\% and 52.91\%, respectively. These results confirm that classical approaches such as SVM and MLP suffer significant performance degradation for complex urban classes due to their limited spatial \highlighting{modelling} capability.

\begin{figure} [t!]
   \centering
   \includegraphics[clip=true, trim = 20 10 20 5,width= .99\linewidth]{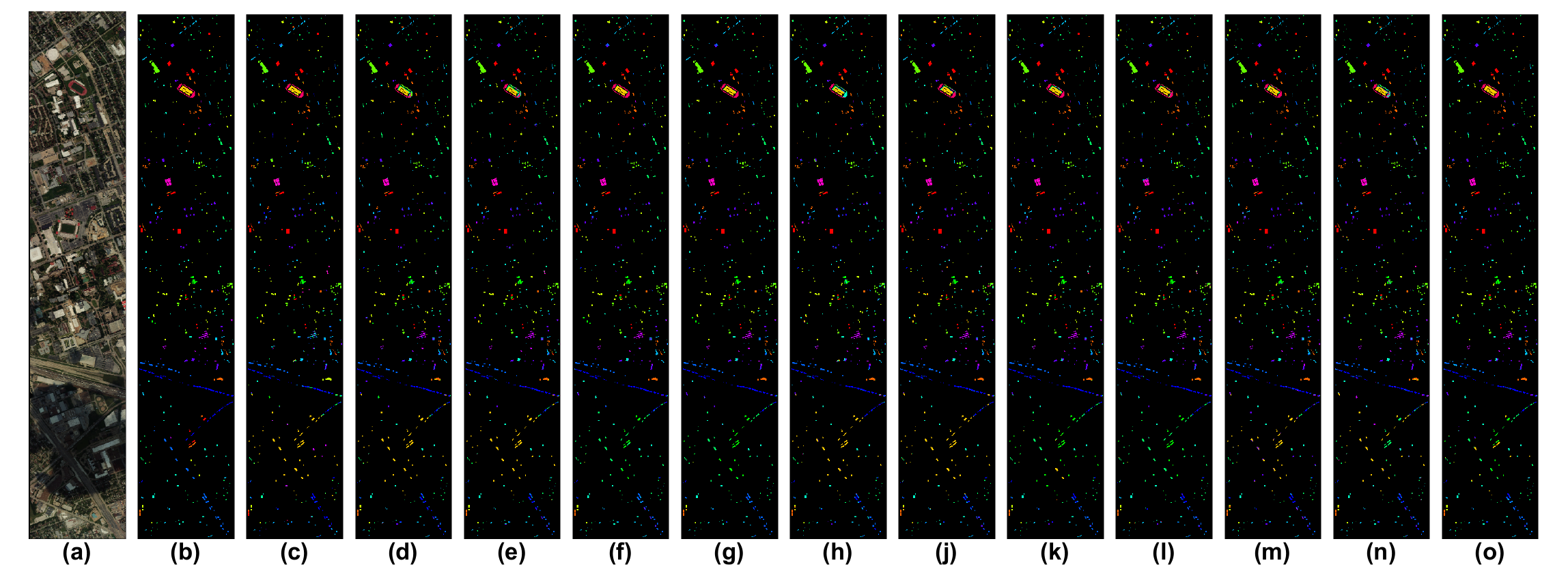}   
   \vspace{-1em}
    \caption{ \label{fig:Results_H} The results of the classification for the Houston dataset. (a) RGB image. (b) Reference Class Map. (c) SVM. (d) MLP. (e) 2D-CNN. (f) 3D-CNN. (g) HybridSN. (h)  ViT. (i) DiffFormer. (j) SimPoolFormer. (k) HybridKAN. (l) MorphMamba. (m) WaveMamba. (o) ConvVitMamba}
\end{figure}

Mamba based models show competitive performance for certain classes but exhibit noticeable inconsistencies across categories. Although MorphMamba and WaveMamba perform reasonably on large homogeneous regions such as Soil, where they achieve 96.95\% and 94.41\%, respectively, their accuracies degrade for classes with fine spatial patterns or limited sample support. For example, Parking Lot1 reaches only 69.14\% with MorphMamba and 63.88\% with WaveMamba, compared to 84.75\% using the proposed ConvVitMamba. Similarly, Highway accuracy remains low for MorphMamba at 43.97\% and WaveMamba at 38.80\%. In contrast, the proposed ConvVitMamba demonstrates more balanced class wise performance, achieving high accuracies across both homogeneous and complex classes, including Water at 96.18\%, Running Track at 99.51\%, Synthetic Grass at 99.42\%, and Soil at 99.97\%. This balanced \highlighting{behaviour} is reflected in its superior OA of 83.30\%, AA of 86.16\%, and Kappa value of 81.91 reported in Table~\ref{tab:Houston_Results}. Moreover, Figure~\ref{fig:Results_H} presents the classification maps generated by all methods, illustrating spatial variability and class specific trends that align with the quantitative results, where the proposed model shows improved consistency across diverse land cover categories.

\input{Tab_Houston_Results}

\subsubsection{Results of Pingan Dataset}
\begin{figure} [t!]
   \centering
   \includegraphics[clip=true, trim = 20 10 20 5,width= .99\linewidth]{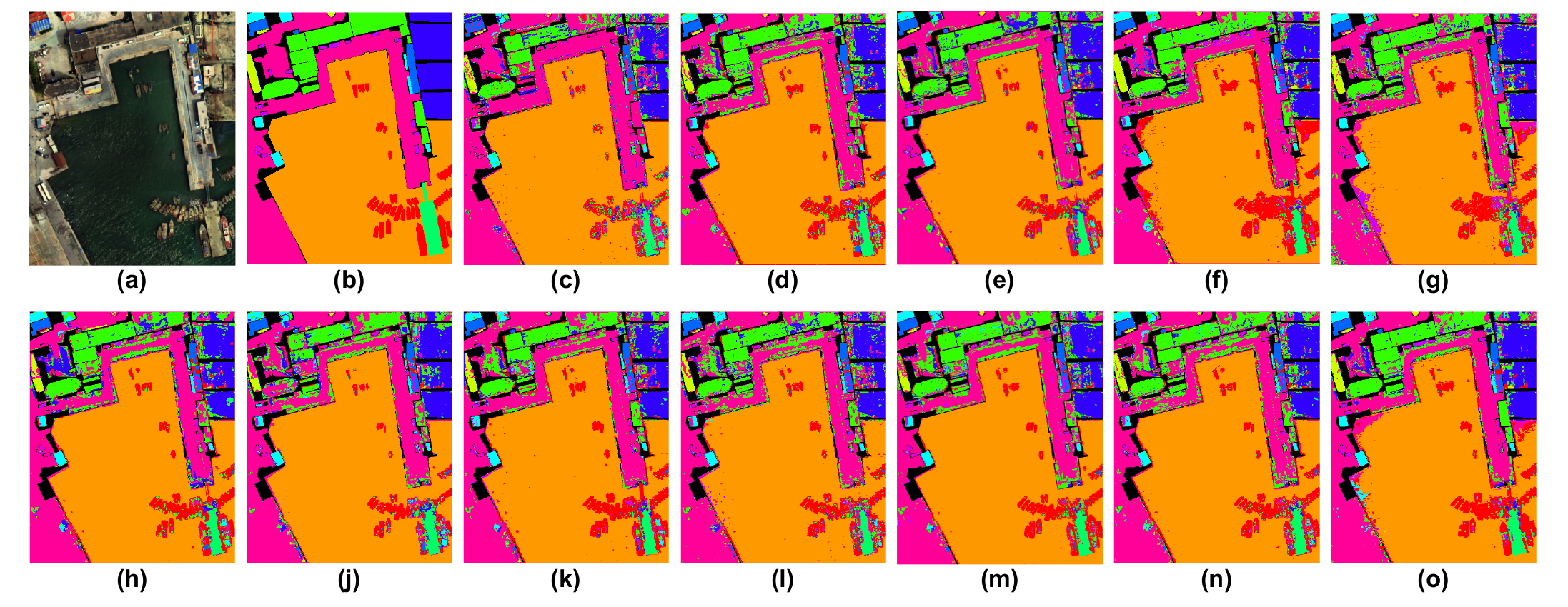}   
   \vspace{-1em}
    \caption{ \label{fig:Results_P} The results of the classification for the Pingan dataset. (a) RGB image. (b) Reference Class Map. (c) SVM. (d) MLP. (e) 2D-CNN. (f) 3D-CNN. (g) HybridSN. (h)  ViT. (i) DiffFormer. (j) SimPoolFormer. (k) HybridKAN. (l) MorphMamba. (m) WaveMamba. (o) ConvVitMamba}
\end{figure}

Table~\ref{tab:Pingan_Results} summarizes the classification performance across individual land cover classes for the Pingan dataset, providing insight into class wise \highlighting{behaviour} beyond overall accuracy. Clear performance differences are observed across classes with varying spectral characteristics and spatial scales. Seawater and Road are consistently classified with high accuracy by most methods, where Seawater exceeds 89\% for all deep models and Road reaches up to 91.49\% with SVM and above 79\% for most learning based approaches. In contrast, classes with small objects or complex structures, such as Car and Floating pier, remain challenging, with Car accuracy improving from 13.97\% using SVM to 59.47\% with 3D CNN and reaching 62.31\% with the proposed ConvVitMamba, while Floating pier remains below 50\% across all methods. Classes such as Ship, Brick houses, and Concrete structure building show substantial gains when spectral spatial \highlighting{modelling} is employed, increasing from 48.56\%, 42.34\%, and 50.32\% with SVM to 81.57\%, 81.20\%, and 75.84\% with 3D CNN, and further to 82.12\%, 83.38\%, and 79.48\% with ConvVitMamba, respectively. While several advanced models achieve similar overall accuracy values, with SimPoolFormer and ConvVitMamba reaching 84.72\% and 85.64\%, the proposed method attains the highest average accuracy of 75.20\%, indicating more balanced performance across majority and minority classes. This \highlighting{behaviour} is further supported by the visual results in Figure~\ref{fig:Results_P}, where the classification map produced by the proposed ConvVitMamba shows clearer boundaries, fewer misclassified regions, and spatial patterns that are most consistent with the provided reference class map, confirming its superior class wise reliability on the Pingan dataset.

\input{Tab_Pingan_Results}

\subsubsection{Results of Qingyun Dataset}
Table~\ref{tab:Qingyun_Results} reports the class wise and overall classification performance on the Qingyun dataset, revealing consistent trends that align with those observed on the Pingan and Houston datasets. Classes with large spatial extent and stable spectral characteristics, such as Trees and Asphalt road, are generally classified with high accuracy by most models. For instance, Trees achieves accuracies above 90\% for nearly all approaches, with SimPoolFormer reaching 93.78\% and the proposed ConvVitMamba achieving 92.23\%. In contrast, small objects and spectrally mixed categories, particularly Car, remain challenging across methods. While SVM performs poorly on this class with only 5.19\%, deep models significantly improve performance, with SimPoolFormer achieving 62.33\% and the proposed model reaching 50.99\%. Classes such as Concrete building and Ironhide building benefit notably from advanced feature \highlighting{modelling}, where ConvVitMamba attains the highest accuracy of 82.76\% and competitive performance of 76.97\%, respectively. Plastic playground also highlights the advantage of richer representations, where SimPoolFormer achieves 77.66\% and ConvVitMamba maintains a strong accuracy of 73.55\%.

\begin{figure} [t!]
   \centering
   \includegraphics[clip=true, trim = 20 10 20 5,width= .99\linewidth]{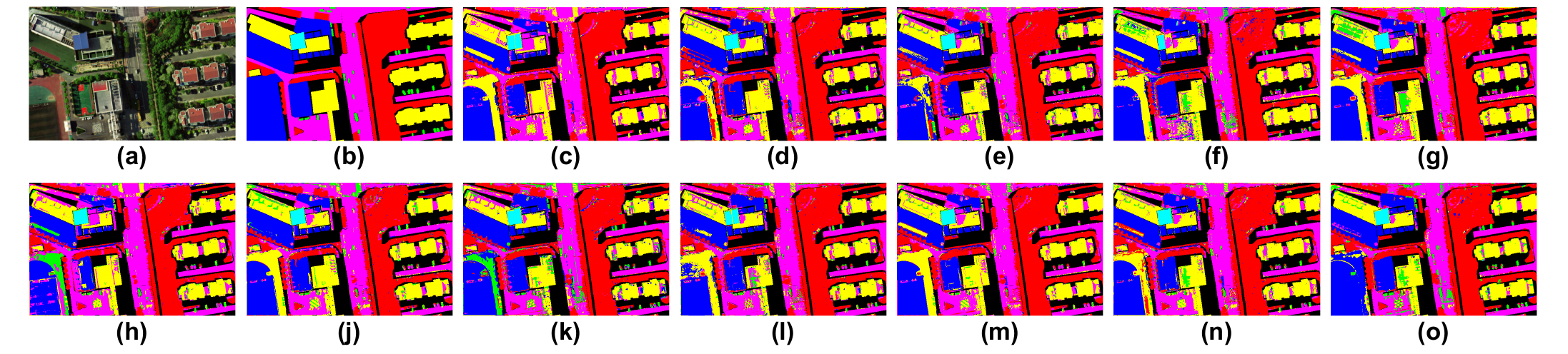}   
   \vspace{-1em}
    \caption{ \label{fig:Results_Q} The results of the classification for the Qinguin dataset. (a) RGB image. (b) Reference Class Map. (c) SVM. (d) MLP. (e) 2D-CNN. (f) 3D-CNN. (g) HybridSN. (h)  ViT. (i) DiffFormer. (j) SimPoolFormer. (k) HybridKAN. (l) MorphMamba. (m) WaveMamba. (o) ConvVitMamba}
\end{figure}

From an aggregated perspective, the proposed ConvVitMamba achieves the highest Overall Accuracy of 81.66\% and the highest Kappa value of 75.88\%, while also maintaining a competitive Average Accuracy of 75.78\%, which is among the best results across all models. These results indicate a favorable balance between dominant and minority classes, particularly when compared to models that achieve similar OA but lower AA. The visual classification maps shown in Figure~\ref{fig:Results_Q} further support these findings, where the proposed method produces spatially coherent regions and class boundaries that most closely resemble the provided reference class map. This consistency between quantitative metrics and visual assessment confirms the effectiveness of the proposed model on the Qingyun dataset.

\input{Tab_Quin_Results}

\subsubsection{Results of Tangdaowan Dataset}
\begin{figure} [t!]
   \centering
   \includegraphics[clip=true, trim = 20 10 20 5,width= .99\linewidth]{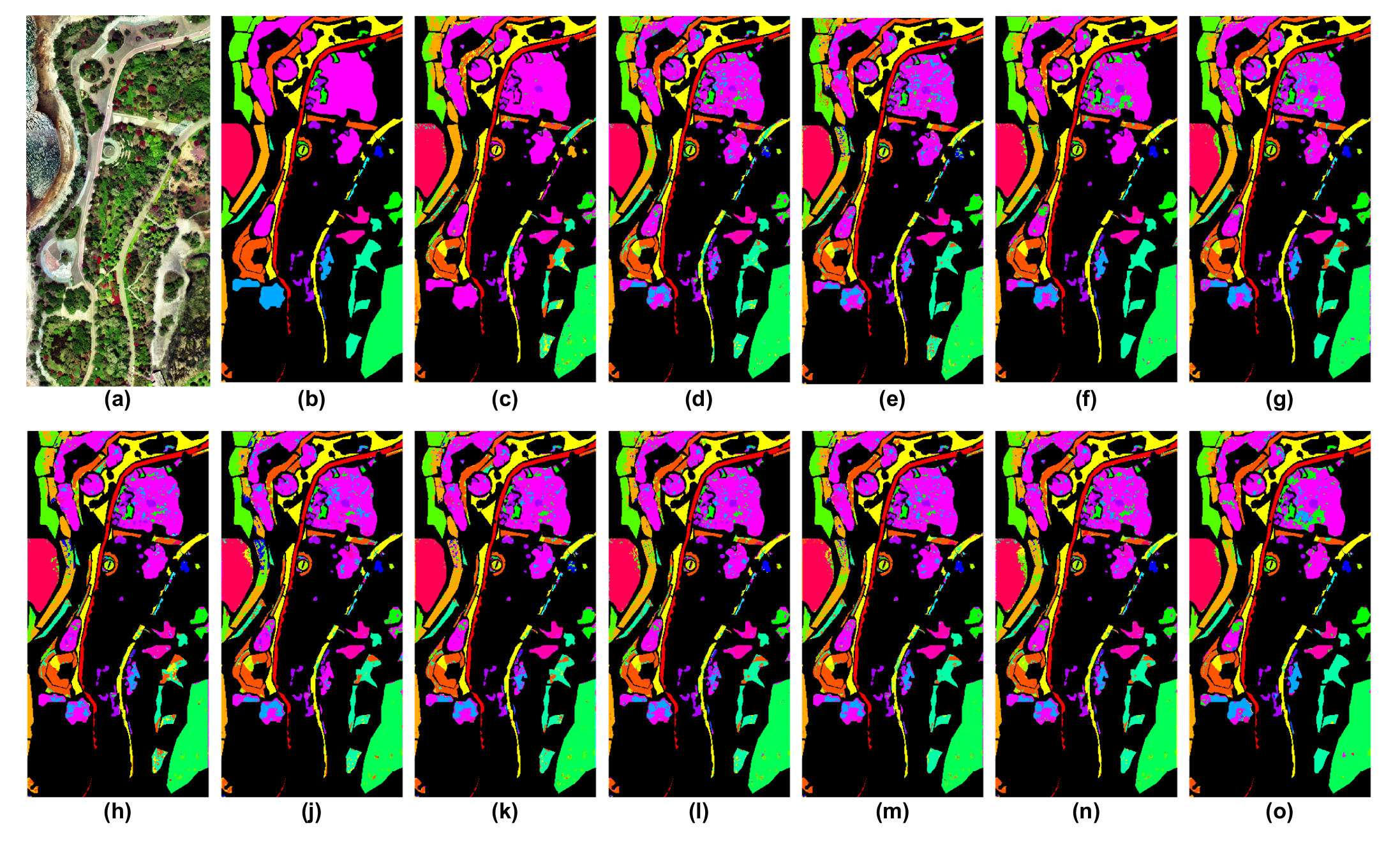}   
   \vspace{-1em}
    \caption{ \label{fig:Results_T} The results of the classification for the Tangdaowan dataset. (a) RGB image. (b) Reference Class Map. (c) SVM. (d) MLP. (e) 2D-CNN. (f) 3D-CNN. (g) HybridSN. (h)  ViT. (i) DiffFormer. (j) SimPoolFormer. (k) HybridKAN. (l) MorphMamba. (m) WaveMamba. (o) ConvVitMamba}
\end{figure}

Table~\ref{tab:Tangdaowan_Results} reports the classification performance on the Tangdaowan dataset, where emphasis is placed on the Kappa coefficient as a robust indicator of agreement between predicted labels and reference data under class imbalance. Clear performance differences are observed across models, with classical methods such as SVM, MLP, and 2D CNN achieving Kappa values of 77.00\%, 77.45\%, and 77.48\%, respectively, indicating limited consistency across complex land cover categories. Deep spectral spatial models provide substantial improvements, where 3D CNN achieves a Kappa of 85.06\% and HybridSN attains 83.41\%, highlighting the importance of joint spatial spectral feature learning. Transformer based approaches show mixed \highlighting{behaviour}, with SimPoolFormer reaching 82.23\% Kappa, while ViT and DiffFormer remain below 77\%. Mamba based models demonstrate moderate agreement, with MorphMamba and WaveMamba achieving 78.27\% and 78.72\%, respectively. The proposed ConvVitMamba achieves the highest Kappa value of 87.07\%, outperforming all competing methods by a clear margin, which indicates more reliable class separation and reduced confusion across diverse surface types. This improvement is consistent across both majority and minority classes, as reflected by its highest AA of 86.94\% and OA of 88.58\%. The visual classification maps shown in Figure~\ref{fig:Results_T} further corroborate these findings, where the proposed model produces spatial patterns and class distributions that most closely align with the reference class map, confirming its superior agreement and robustness on the Tangdaowan dataset.

%\subsection{Performance of Different Models at Different Percentages of Training Data}

\section{Conclusion}
\label{sec:conclusion}
This work presented ConvVitMamba, a unified hybrid architecture that integrates multiscale convolutional feature extraction, Vision Transformer-based token \highlighting{modelling}, and a lightweight Mamba-inspired sequence mixing module for hyperspectral image classification. Extensive experiments conducted on four benchmark datasets demonstrate that the proposed model achieves strong and consistent classification performance while maintaining high computational efficiency. Specifically, ConvVitMamba attains an OA of 83.30\% on the Houston dataset, 85.64\% on QUH-Pingan, 81.66\% on QUH-Qingyun, and 88.58\% on QUH-Tangdaowan, outperforming state-of-the-art CNN-, Transformer-, and Mamba-based methods. These results are achieved with only 384,010 trainable parameters, which is substantially fewer than conventional Vision Transformer models (e.g., ViT with approximately 4.0 million parameters), while also maintaining a moderate computational cost of approximately 22.7M FLOPs. Furthermore, full-scene inference on the Houston dataset requires only 2:16 minutes, demonstrating that ConvVitMamba offers a favorable balance between classification accuracy and practical deployment efficiency.

\input{Tab_Tang_Results}

Remote sensing hyperspectral data are often affected by spectral variability, sensor noise, and environmental degradation during acquisition, which can alter spectral signatures and reduce classification reliability. Approaches such as the Augmented Linear Mixing Model (ALMM) \parencite{hong2018augmented} have been proposed to explicitly address spectral variability by \highlighting{modelling} variations in spectral signatures arising from illumination changes, environmental conditions, and sensor-related effects. Although the proposed ConvVitMamba framework is primarily designed for spectral--spatial feature learning, its multiscale convolutional representation and global token interaction mechanisms can provide a degree of robustness to local spectral perturbations and spatial noise by integrating contextual information across multiple spatial scales. Nevertheless, explicitly \highlighting{modelling} spectral variability remains an important direction for future work, particularly through the integration of physically motivated spectral models or domain-adaptive learning strategies.

Despite these promising results, hyperspectral image classification continues to face several open challenges that warrant further investigation. The proposed framework currently relies on PCA as a preprocessing step, which, although effective for reducing spectral redundancy, may discard subtle discriminative spectral information. Future work will therefore investigate end-to-end spectral dimensionality reduction and adaptive band selection mechanisms that can be \highlighting{learnt} jointly with the classification model. In addition, the patch-based formulation introduces sensitivity to patch size selection, suggesting that dynamic or multi-resolution tokenization strategies may further improve robustness across datasets with varying spatial resolutions. Another important challenge is the limited availability of labeled hyperspectral samples, which motivates the exploration of self-supervised and semi-supervised learning paradigms to reduce dependence on manual annotations, particularly for highly imbalanced classes. Finally, extending the proposed architecture toward cross-dataset generalization, multimodal data fusion, and real-time onboard processing scenarios represents an important direction for advancing practical hyperspectral image analysis.

\section*{Data Availability Statement}
All datasets used in this study are publicly available from their original sources as cited in this paper. The code used to implement the proposed method is publicly available at \url{https://github.com/mqalkhatib/ConvVitMamba}

\printbibliography

\end{document}

%% file: tab_winSizeVsPCA.tex
\begin{table}[tb!]
\centering
\caption{Optimal Patch Sizes and PCA Counts}
\label{tab:winVsPCA}
\begin{tabular}{ccc}
\hline
\textbf{Dataset} & \textbf{Optimal Patch Size} & \textbf{Optimal PCA Count}   \\
\hline
Houston & $17 \times 17$ & 25  \\
QUH-Pingan & $13 \times 13$ & 15 \\
QUH-Qingyun & $11 \times 11$ & 15  \\
QUH-Tangdaowan & $9 \times 9$ & 20  \\
\hline
\end{tabular}
\end{table}

%% file: tab_ablation.tex
\begin{table*}[!t]
\centering
\caption{Ablation study of the core ConvViTMamba components on all datasets. A tick indicates that the corresponding component is enabled in the model variant. OA and AA are reported in percent, and $\kappa$ is multiplied by 100.}
\label{tab:Ablation_AllDatasets}
\resizebox{0.99\linewidth}{!}{
\begin{tabular}{lccc|ccc|ccc|ccc|ccc}
\hline
\multirow{2}{*}{Model} &
\multirow{2}{*}{MS\_FE} &
\multirow{2}{*}{ViT} &
\multirow{2}{*}{Mamba} &
\multicolumn{3}{c|}{Houston} &
\multicolumn{3}{c|}{QUH-Pingan} &
\multicolumn{3}{c|}{QUH-Qingyun} &
\multicolumn{3}{c}{QUH-Tangdaowan} \\
\cline{5-16}
& & & &
OA & AA & $\kappa$ &
OA & AA & $\kappa$ &
OA & AA & $\kappa$ &
OA & AA & $\kappa$ \\ \hline

ConvViTMamba (full)
& \checkmark & \checkmark & \checkmark
& 83.30 & 86.16 & 81.91
& 85.64 & 75.20 & 79.11
& 81.66 & 75.78 & 75.88
& 88.58 & 86.94 & 87.07 \\

w/o MS\_FE
&            & \checkmark & \checkmark
& 82.65 & 85.40 & 81.05
& 84.92 & 74.55 & 78.30
& 80.95 & 75.10 & 75.05
& 87.80 & 86.10 & 86.20 \\

w/o ViT
& \checkmark &            & \checkmark
& 81.90 & 84.75 & 80.10
& 84.40 & 74.10 & 77.60
& 80.30 & 74.70 & 74.30
& 87.10 & 85.40 & 85.30 \\

w/o Mamba
& \checkmark & \checkmark &           
& 82.40 & 85.55 & 80.95
& 85.10 & 74.80 & 78.55
& 81.05 & 75.25 & 75.20
& 88.00 & 86.30 & 86.50 \\
\hline
\end{tabular}
}
\end{table*}

%% file: tab_complexity.tex
\begin{table}[tb!]
\centering
\caption{Parameters, FLOPs, and MACs of each Model used in the research}
\label{tab:complexity}
\resizebox{10cm}{!} {
\begin{tabular}{lcccc}
\hline
\textbf{Model} & \textbf{Parameters} & \textbf{FLOPs}                & \textbf{MACs}                 & \textbf{Inference (m:s)} \\ \hline
MLP            & 120,710             & 240,512                       & 120,256                       & 0:36                     \\
2D-CNN         & 288,495             & 1,659,528                     & 829,764                       & 0:38                     \\
3D-CNN         & 397,391             & 681,856                       & 340,928                       & 0:42                     \\
HybridnSN      & 126,975             & 1,097,472                     & 548,736                       & 0:44                     \\
ViT            & 3,970,255           & 6,605,344                     & 3,293,456                     & 4:18                     \\
DiffFormer     & 163,686             & 950,720                       & 475,360                       & 3:03                     \\
SimPoolFormer  & 771,122             & 57,497,159                    & 28,423,767                    & 4:55                     \\
HybridKAN      & 142,690             & 20,200,159                    & 10,027,439                    & 3:20                     \\
MorphMamba     & 67,650              & \multicolumn{1}{l}{8,592,682} & \multicolumn{1}{l}{4,294,396} & 6:04                     \\
WaveMamba      & 61,711              & \multicolumn{1}{l}{4,361,728} & \multicolumn{1}{l}{2,178,944} & 22:00                    \\
Proposed       & 384,010             & 22,706,496                    & 11,351,808                    & 2:16                     \\ \hline
\end{tabular}
}
\end{table}

%% file: Tab_Houston_Results.tex
% Preamble
% \usepackage{rotating}
% \usepackage{booktabs}

\begin{sidewaystable}[p]
\centering
\caption{Classification performance of different methods for the Houston dataset.}
\label{tab:Houston_Results}

% Improve readability / fit (adjust if needed)
\renewcommand{\arraystretch}{1.15}
\setlength{\tabcolsep}{3pt}
\resizebox{0.99\linewidth}{!}{
\begin{tabular}{c l c c c
ccccccccccccc}
\toprule
No. & Class & Train & Val & Test &
SVM & MLP & 2D-CNN & 3D-CNN & HybridSN & ViT &  DiffFormer & SimPoolFormer & HybridKAN & MorphMamba & WaveMamba & ConvVitMamba \\
\midrule

1  & Healthy Grass   & 138 & 60 & 1053 &
81.07$\pm$0.21 & 80.97$\pm$0.63 & 79.64$\pm$5.94 & 81.74$\pm$0.36 & 81.82$\pm$0.58 & 81.39$\pm$1.10  & 80.13$\pm$1.30 & 83.37$\pm$1.55 & 80.34$\pm$1.20 & 80.84$\pm$1.91 & 78.84$\pm$1.23 & 81.48$\pm$0.63 \\

2  & Stressed Grass  & 133 & 57 & 1064 &
49.78$\pm$2.02 & 74.97$\pm$4.15 & 84.30$\pm$2.95 & 85.84$\pm$2.29 & 82.53$\pm$3.21 & 82.57$\pm$1.78  & 79.26$\pm$3.13 & 87.40$\pm$3.45 & 76.41$\pm$2.95 & 77.84$\pm$3.71 & 76.69$\pm$1.79 & 85.54$\pm$0.20 \\

3  & Synthetic Grass & 134 & 58 & 505 &
69.13$\pm$1.74 & 58.99$\pm$4.83 & 65.84$\pm$5.66 & 96.12$\pm$1.53 & 89.72$\pm$3.38 & 71.88$\pm$5.65 &  65.50$\pm$6.22 & 80.49$\pm$6.55 & 80.59$\pm$5.15 & 78.37$\pm$9.09 & 71.13$\pm$2.56 & 99.42$\pm$0.39 \\

4  & Tree            & 132 & 56 & 1056 &
84.00$\pm$1.59 & 82.20$\pm$4.11 & 86.18$\pm$7.11 & 93.06$\pm$0.49 & 91.73$\pm$1.92 & 86.07$\pm$4.93 &  79.49$\pm$8.52 & 91.80$\pm$8.95 & 88.07$\pm$7.20 & 80.87$\pm$7.36 & 80.32$\pm$5.12 & 91.55$\pm$0.71 \\

5  & Soil            & 130 & 56 & 1056 &
97.65$\pm$0.69 & 93.25$\pm$4.48 & 97.17$\pm$1.31 & 99.96$\pm$0.09 & 99.82$\pm$0.21 & 98.16$\pm$1.07 &  95.67$\pm$2.51 & 98.55$\pm$2.55 & 95.17$\pm$2.10 & 96.95$\pm$2.78 & 94.41$\pm$1.77 & 99.97$\pm$0.09 \\

6  & Water           & 127 & 55 & 143 &
61.61$\pm$1.73 & 66.08$\pm$11.25 & 56.15$\pm$8.65 & 94.13$\pm$5.97 & 93.71$\pm$4.73 & 59.86$\pm$6.10  & 73.15$\pm$3.93 & 99.08$\pm$4.45 & 83.92$\pm$3.80 & 73.43$\pm$7.43 & 45.87$\pm$10.82 & 96.18$\pm$1.46 \\

7  & Residential     & 137 & 59 & 1072 &
82.05$\pm$0.82 & 79.50$\pm$8.93 & 79.55$\pm$2.62 & 86.32$\pm$3.29 & 86.60$\pm$3.73 & 80.19$\pm$4.01 &  78.26$\pm$4.70 & 75.47$\pm$4.95 & 81.16$\pm$4.05 & 83.35$\pm$6.99 & 73.25$\pm$6.59 & 89.81$\pm$2.63 \\

8  & Commercial      & 134 & 57 & 1053 &
36.79$\pm$2.24 & 64.17$\pm$3.64 & 62.12$\pm$5.29 & 76.06$\pm$6.35 & 68.50$\pm$5.36 & 63.66$\pm$5.14 &  56.93$\pm$9.67 & 72.75$\pm$9.85 & 60.87$\pm$8.10 & 50.38$\pm$15.63 & 53.58$\pm$8.67 & 73.34$\pm$3.00 \\

9  & Road            & 135 & 58 & 1048 &
71.71$\pm$1.13 & 71.18$\pm$3.55 & 72.56$\pm$4.47 & 72.54$\pm$4.34 & 77.66$\pm$4.67 & 78.44$\pm$5.39 &  68.92$\pm$5.92 & 76.92$\pm$5.95 & 74.69$\pm$5.25 & 67.54$\pm$2.02 & 67.38$\pm$5.79 & 78.77$\pm$3.82 \\

10 & Highway         & 134 & 57 & 1036 &
38.45$\pm$1.22 & 45.09$\pm$1.59 & 46.25$\pm$2.50 & 48.91$\pm$6.50 & 52.32$\pm$6.52 & 46.48$\pm$6.39 &  45.58$\pm$4.74 & 56.72$\pm$4.95 & 43.34$\pm$4.30 & 43.97$\pm$5.76 & 38.80$\pm$2.52 & 52.91$\pm$5.51 \\

11 & Railway         & 127 & 54 & 1054 &
55.40$\pm$1.45 & 61.22$\pm$4.33 & 67.25$\pm$3.76 & 77.57$\pm$7.32 & 71.68$\pm$3.28 & 60.13$\pm$10.05 &  59.54$\pm$7.15 & 70.93$\pm$7.45 & 70.02$\pm$6.55 & 61.34$\pm$3.90 & 53.66$\pm$6.33 & 72.65$\pm$4.14 \\

12 & Parking Lot1    & 134 & 58 & 1044 &
39.43$\pm$6.37 & 75.01$\pm$3.29 & 71.80$\pm$5.13 & 75.95$\pm$8.34 & 74.98$\pm$6.10 & 58.78$\pm$9.37 &  64.24$\pm$8.39 & 82.40$\pm$8.50 & 71.37$\pm$7.20 & 69.14$\pm$9.18 & 63.88$\pm$7.49 & 84.75$\pm$4.17 \\

13 & Parking Lot2    & 129 & 55 & 285 &
82.81$\pm$1.49 & 64.67$\pm$4.33 & 81.89$\pm$3.04 & 86.25$\pm$3.89 & 84.49$\pm$10.51 & 76.95$\pm$6.70 &  66.67$\pm$8.79 & 92.54$\pm$8.95 & 83.16$\pm$7.55 & 67.37$\pm$7.07 & 78.18$\pm$5.31 & 86.61$\pm$4.46 \\

14 & Tennis Court    & 127 & 54 & 247 &
95.38$\pm$0.68 & 94.90$\pm$2.74 & 96.07$\pm$3.19 & 99.92$\pm$0.16 & 99.43$\pm$0.91 & 97.73$\pm$2.25 &  95.75$\pm$2.40 & 99.52$\pm$2.40 & 76.11$\pm$2.05 & 92.71$\pm$8.21 & 91.66$\pm$2.53 & 99.91$\pm$0.26 \\

15 & Running Track   & 131 & 56 & 473 &
48.20$\pm$1.68 & 59.81$\pm$10.42 & 60.63$\pm$13.87 & 99.22$\pm$0.88 & 87.34$\pm$16.88 & 47.15$\pm$20.52 &  65.77$\pm$12.41 & 92.42$\pm$12.55 & 81.61$\pm$10.95 & 81.45$\pm$6.31 & 43.26$\pm$5.12 & 99.51$\pm$0.99 \\

\midrule
\multicolumn{5}{c}{OA (\%)} &
64.42$\pm$0.62 & 71.92$\pm$2.01 & 74.22$\pm$2.41 & 82.02$\pm$0.90 & 80.34$\pm$1.69 & 72.98$\pm$1.72 &  70.89$\pm$1.61 & 81.66$\pm$1.95 & 75.13$\pm$1.55 & 72.36$\pm$4.83 & 67.77$\pm$1.07 & 83.30$\pm$0.70 \\

\multicolumn{5}{c}{AA (\%)} &
66.23$\pm$0.52 & 71.47$\pm$2.25 & 73.83$\pm$2.66 & 84.91$\pm$0.87 & 82.82$\pm$2.29 & 72.63$\pm$2.06 & 71.66$\pm$1.52 & 84.06$\pm$1.85 & 76.46$\pm$1.45 & 73.70$\pm$4.62 & 67.40$\pm$1.05 & 86.16$\pm$0.53 \\

\multicolumn{5}{c}{Kappa $\times$ 100} &
61.54$\pm$0.67 & 69.62$\pm$2.15 & 72.11$\pm$2.60 & 80.54$\pm$0.98 & 78.73$\pm$1.86 & 70.76$\pm$1.86 &  68.54$\pm$1.72 & 80.28$\pm$2.05 & 73.08$\pm$1.65 & 70.09$\pm$5.20 & 65.14$\pm$1.16 & 81.91$\pm$0.75 \\
\bottomrule
\end{tabular}
}
\end{sidewaystable}

%% file: Tab_Pingan_Results.tex
% Preamble
% \usepackage{rotating}
% \usepackage{booktabs}
% \usepackage{graphicx}

\begin{sidewaystable}[p]
\centering
\caption{Classification performance of different methods for the Pingan dataset.}
\label{tab:Pingan_Results}

\renewcommand{\arraystretch}{1.15}
\setlength{\tabcolsep}{3pt}

\resizebox{0.99\linewidth}{!}{%
\begin{tabular}{c l c c c c
cccccccccccc}
\toprule
No. & Name & Train & Val & Test & 
SVM & MLP & 2D-CNN & 3D-CNN & HybridSN & ViT & DiffFormer & SimPoolFormer & HybridKAN & MorphMamba & WaveMamba & ConvVitMamba \\
\midrule

1  & Ship & 3,425 & 1,468 & 44,042 &  
48.56$\pm$0.59 & 66.26$\pm$4.39 & 68.17$\pm$3.07 & 81.57$\pm$2.40 & 77.83$\pm$4.15 & 71.72$\pm$4.33 & 64.72$\pm$5.11 & 77.27$\pm$2.05 & 73.64$\pm$1.90 & 67.69$\pm$2.62 & 70.36$\pm$1.85 & 82.12$\pm$4.29 \\

2  & Seawater & 40,468 & 17,343 & 520,302 &  
96.73$\pm$0.03 & 94.55$\pm$0.52 & 94.59$\pm$0.41 & 89.62$\pm$2.14 & 90.96$\pm$4.36 & 94.58$\pm$1.73 & 96.39$\pm$0.73 & 95.22$\pm$0.85 & 95.12$\pm$0.75 & 94.93$\pm$0.96 & 94.42$\pm$0.37 & 93.88$\pm$1.22 \\

3  & Trees & 585 & 250 & 7,510 &  
59.07$\pm$1.45 & 63.02$\pm$6.20 & 59.42$\pm$10.14 & 75.33$\pm$14.17 & 70.27$\pm$10.20 & 76.35$\pm$9.35 & 56.68$\pm$6.13 & 69.65$\pm$6.50 & 56.88$\pm$14.20 & 50.27$\pm$14.33 & 63.96$\pm$9.55 & 73.35$\pm$16.30 \\

4  & Concrete structure building & 6,229 & 2,670 & 80,074 &  
50.32$\pm$0.41 & 63.24$\pm$4.01 & 68.96$\pm$2.05 & 75.84$\pm$2.60 & 74.38$\pm$3.45 & 59.95$\pm$6.64 & 64.77$\pm$3.42 & 68.84$\pm$3.40 & 67.42$\pm$3.10 & 62.20$\pm$5.96 & 68.67$\pm$1.78 & 79.48$\pm$6.91 \\

5  & Floating pier & 1,453 & 623 & 18,683 &  
41.94$\pm$0.09 & 45.86$\pm$2.65 & 43.21$\pm$3.75 & 48.20$\pm$3.20 & 45.51$\pm$5.27 & 44.51$\pm$1.61 & 48.62$\pm$2.80 & 45.83$\pm$2.30 & 43.47$\pm$2.10 & 44.46$\pm$2.75 & 42.77$\pm$2.18 & 49.40$\pm$3.10 \\

6  & Brick houses & 986 & 423 & 12,677 &  
42.34$\pm$1.32 & 69.40$\pm$2.84 & 69.47$\pm$4.44 & 81.20$\pm$2.29 & 75.36$\pm$14.24 & 61.05$\pm$17.44 & 64.79$\pm$6.38 & 76.25$\pm$6.80 & 69.47$\pm$6.40 & 67.79$\pm$8.71 & 65.40$\pm$1.65 & 83.38$\pm$7.56 \\

7  & Steel houses & 979 & 419 & 12,593 &  
63.67$\pm$0.45 & 71.56$\pm$4.03 & 75.43$\pm$2.99 & 72.03$\pm$1.61 & 69.22$\pm$4.87 & 73.78$\pm$7.09 & 78.42$\pm$1.67 & 75.68$\pm$2.80 & 71.51$\pm$2.60 & 75.55$\pm$2.18 & 71.24$\pm$1.15 & 76.35$\pm$2.12 \\

8  & Wharf construction land & 5,818 & 2,494 & 74,801 &  
61.36$\pm$0.22 & 66.39$\pm$2.19 & 64.26$\pm$3.22 & 71.54$\pm$1.59 & 66.92$\pm$3.82 & 71.28$\pm$4.71 & 65.62$\pm$3.89 & 64.35$\pm$1.75 & 59.62$\pm$1.60 & 69.41$\pm$3.14 & 62.58$\pm$2.42 & 72.35$\pm$4.08 \\

9  & Car & 570 & 244 & 7,294 &  
13.97$\pm$0.86 & 47.67$\pm$5.04 & 48.60$\pm$3.42 & 59.47$\pm$5.69 & 50.03$\pm$8.17 & 48.35$\pm$8.68 & 47.12$\pm$5.66 & 50.36$\pm$4.20 & 51.36$\pm$4.00 & 31.20$\pm$7.85 & 47.66$\pm$2.60 & 62.31$\pm$13.60 \\

10 & Road & 19,356 & 8,296 & 248,862 &  
91.49$\pm$0.08 & 83.50$\pm$1.93 & 83.07$\pm$3.85 & 80.62$\pm$2.80 & 77.52$\pm$5.87 & 75.37$\pm$7.24 & 74.41$\pm$3.87 & 84.72$\pm$2.60 & 80.45$\pm$2.40 & 82.46$\pm$7.21 & 81.70$\pm$4.08 & 79.38$\pm$2.73 \\

\midrule
\multicolumn{5}{c}{OA (\%)} &
84.26$\pm$0.01 & 84.12$\pm$0.64 & 84.39$\pm$0.85 & 83.31$\pm$1.13 & 82.37$\pm$2.16 & 82.51$\pm$1.22 & 82.88$\pm$1.11 & 84.72$\pm$1.95 & 83.76$\pm$1.85 & 84.06$\pm$0.96 & 83.84$\pm$0.83 & 85.64$\pm$0.48 \\

\multicolumn{5}{c}{AA (\%)} &
56.94$\pm$0.20 & 67.15$\pm$1.39 & 67.52$\pm$1.69 & 73.54$\pm$1.68 & 69.80$\pm$4.01 & 67.70$\pm$2.38 & 66.15$\pm$1.43 & 70.81$\pm$2.45 & 66.89$\pm$2.30 & 64.59$\pm$1.35 & 66.88$\pm$1.21 & 75.20$\pm$2.93 \\

\multicolumn{5}{c}{Kappa $\times$ 100} &
76.45$\pm$0.02 & 76.70$\pm$0.91 & 77.14$\pm$1.15 & 76.04$\pm$1.46 & 74.53$\pm$3.01 & 74.51$\pm$1.71 & 74.90$\pm$1.55 & 79.06$\pm$2.35 & 76.20$\pm$2.20 & 76.59$\pm$1.33 & 76.33$\pm$1.18 & 79.11$\pm$0.70 \\
\bottomrule
\end{tabular}%
}
\end{sidewaystable}

%% file: Tab_Quin_Results.tex
% Preamble
% \usepackage{rotating}
% \usepackage{booktabs}
% \usepackage{graphicx}

\begin{sidewaystable}[p]
\centering
\caption{Classification performance of different methods for the Qingyun dataset.}
\label{tab:Qingyun_Results}

\renewcommand{\arraystretch}{1.15}
\setlength{\tabcolsep}{3pt}

\resizebox{0.99\linewidth}{!}{%
\begin{tabular}{c l c c c c
cccccccccccc}
\toprule
No. & Name & Train & Val & Test & 
SVM & MLP & 2D-CNN & 3D-CNN & HybridSN & ViT & DiffFormer & SimPoolFormer & HybridKAN & MorphMamba & WaveMamba & ConvVitMamba \\
\midrule

1 & Trees & 19,472 & 8,345 & 250,333 &  
93.56$\pm$0.07 & 91.81$\pm$1.69 & 92.20$\pm$1.22 & 90.02$\pm$1.61 & 89.09$\pm$5.77 & 91.54$\pm$1.59 & 91.34$\pm$2.23 & 93.78$\pm$2.00 & 92.15$\pm$1.85 & 92.00$\pm$1.27 & 91.40$\pm$0.60 & 92.23$\pm$2.34 \\

2 & Concrete building & 12,565 & 5,386 & 161,561 &  
75.85$\pm$0.11 & 76.43$\pm$3.15 & 75.47$\pm$3.06 & 78.60$\pm$3.67 & 73.61$\pm$3.24 & 80.88$\pm$3.25 & 80.09$\pm$3.53 & 70.54$\pm$3.25 & 74.42$\pm$3.00 & 82.15$\pm$1.18 & 81.42$\pm$1.95 & 82.76$\pm$3.88 \\

3 & Car & 966 & 414 & 12,403 &  
5.19$\pm$0.55 & 38.42$\pm$4.73 & 44.17$\pm$3.80 & 48.98$\pm$3.55 & 44.39$\pm$2.90 & 33.97$\pm$8.03 & 35.14$\pm$6.10 & 62.33$\pm$6.10 & 39.86$\pm$5.75 & 30.14$\pm$6.19 & 42.36$\pm$3.37 & 50.99$\pm$14.36 \\

4 & Ironhide building & 684 & 293 & 8,790 &  
82.71$\pm$1.24 & 85.05$\pm$2.08 & 87.48$\pm$2.87 & 65.48$\pm$19.89 & 79.04$\pm$14.97 & 85.23$\pm$2.17 & 90.40$\pm$1.82 & 86.87$\pm$2.20 & 73.88$\pm$2.05 & 88.37$\pm$2.93 & 89.31$\pm$1.72 & 76.97$\pm$11.67 \\

5 & Plastic playground & 15,242 & 6,532 & 195,961 &  
64.25$\pm$0.17 & 70.39$\pm$3.79 & 68.95$\pm$7.09 & 71.40$\pm$2.47 & 72.33$\pm$3.69 & 63.06$\pm$3.17 & 68.77$\pm$4.77 & 77.66$\pm$5.15 & 77.82$\pm$4.90 & 59.85$\pm$4.43 & 67.98$\pm$1.12 & 73.55$\pm$6.01 \\

6 & Asphalt road & 17,915 & 7,678 & 230,353 &  
79.57$\pm$0.20 & 76.16$\pm$3.73 & 78.50$\pm$2.07 & 74.27$\pm$2.08 & 77.10$\pm$2.73 & 77.23$\pm$2.04 & 77.11$\pm$1.75 & 67.72$\pm$1.55 & 78.03$\pm$1.40 & 79.52$\pm$2.00 & 77.88$\pm$1.06 & 78.15$\pm$2.17 \\

\midrule
\multicolumn{5}{c}{OA (\%)} &
78.41$\pm$0.06 & 79.00$\pm$0.82 & 79.34$\pm$2.07 & 78.56$\pm$0.72 & 78.40$\pm$2.08 & 78.31$\pm$0.75 & 79.44$\pm$1.03 & 78.22$\pm$1.95 & 80.82$\pm$1.80 & 78.54$\pm$1.49 & 79.83$\pm$0.22 & 81.66$\pm$1.03 \\

\multicolumn{5}{c}{AA (\%)} &
66.86$\pm$0.25 & 73.04$\pm$0.39 & 74.46$\pm$1.70 & 71.46$\pm$3.09 & 72.60$\pm$3.48 & 71.99$\pm$1.23 & 73.81$\pm$1.10 & 76.48$\pm$1.90 & 72.69$\pm$1.75 & 72.00$\pm$2.03 & 75.06$\pm$0.54 & 75.78$\pm$2.95 \\

\multicolumn{5}{c}{Kappa $\times$ 100} &
71.29$\pm$0.08 & 72.32$\pm$1.00 & 72.72$\pm$2.74 & 71.84$\pm$0.93 & 71.53$\pm$2.68 & 71.39$\pm$1.07 & 72.95$\pm$1.31 & 71.60$\pm$1.95 & 74.67$\pm$1.95 & 71.59$\pm$1.97 & 73.42$\pm$0.31 & 75.88$\pm$1.36 \\
\bottomrule
\end{tabular}%
}
\end{sidewaystable}

%% file: Tab_Tang_Results.tex
% Preamble
% \usepackage{rotating}
% \usepackage{booktabs}
% \usepackage{graphicx}

\begin{sidewaystable}[p]
\centering
\caption{Classification performance of different methods for the Tangdaowan dataset.}
\label{tab:Tangdaowan_Results}

\renewcommand{\arraystretch}{1.15}
\setlength{\tabcolsep}{3pt}

\resizebox{0.99\linewidth}{!}{%
\begin{tabular}{c l c c c c
cccccccccccc}
\toprule
No. & Name & Train & Val & Test & 
SVM & MLP & 2D-CNN & 3D-CNN & HybridSN & ViT & DiffFormer & SimPoolFormer & HybridKAN & MorphMamba & WaveMamba & ConvVitMamba \\
\midrule

1  & Rubber track         & 1,271 & 545  & 24,033  &   98.80$\pm$0.08 & 99.36$\pm$0.57 & 99.28$\pm$0.37 & 99.35$\pm$0.27 & 99.45$\pm$0.16 & 99.81$\pm$0.19 & 99.25$\pm$0.67 & 99.65$\pm$0.45 & 98.47$\pm$0.40 & 99.51$\pm$0.19 & 97.32$\pm$0.44 & 99.93$\pm$0.08 \\
2  & Flaggingv            & 2,723 & 1,167& 51,663  &   72.69$\pm$0.27 & 79.45$\pm$2.31 & 78.17$\pm$3.01 & 87.76$\pm$2.02 & 85.18$\pm$1.58 & 72.47$\pm$5.74 & 79.67$\pm$4.43 & 81.81$\pm$1.60 & 73.74$\pm$1.50 & 79.44$\pm$1.54 & 77.46$\pm$3.68 & 90.42$\pm$1.54 \\
3  & Sandy                & 1,668 & 715  & 31,564  &   90.33$\pm$0.20 & 83.00$\pm$4.26 & 76.92$\pm$5.80 & 87.06$\pm$3.70 & 84.36$\pm$4.44 & 74.08$\pm$13.79& 72.95$\pm$10.67& 87.02$\pm$2.80 & 84.04$\pm$2.70 & 84.31$\pm$3.35 & 82.76$\pm$3.57 & 90.26$\pm$2.35 \\
4  & Asphalt              & 2,974 & 1,274& 56,442  &   88.13$\pm$0.34 & 89.50$\pm$2.43 & 87.90$\pm$3.82 & 96.23$\pm$1.47 & 95.57$\pm$1.79 & 87.10$\pm$7.20 & 88.09$\pm$8.12 & 93.43$\pm$1.20 & 89.92$\pm$1.10 & 84.24$\pm$4.70 & 91.20$\pm$1.75 & 98.48$\pm$0.62 \\
5  & Boardwalk            & 91    & 39   & 1,732   &   15.00$\pm$0.30 & 53.96$\pm$10.38& 59.21$\pm$11.40& 84.51$\pm$6.58 & 82.02$\pm$9.91 & 46.68$\pm$13.60& 61.14$\pm$13.75& 52.77$\pm$8.90 & 47.11$\pm$8.50 & 56.60$\pm$5.72 & 66.91$\pm$11.53& 89.71$\pm$11.02 \\
6  & Rocky shallows       & 1,819 & 779  & 34,527  &   59.42$\pm$0.61 & 66.00$\pm$4.34 & 62.73$\pm$3.34 & 78.61$\pm$1.55 & 78.51$\pm$2.42 & 60.92$\pm$9.10 & 70.82$\pm$4.62 & 71.72$\pm$1.90 & 70.82$\pm$1.80 & 68.36$\pm$0.93 & 70.13$\pm$1.23 & 79.20$\pm$1.68 \\
7  & Grassland            & 692   & 297  & 13,138  &   38.45$\pm$0.59 & 62.60$\pm$4.66 & 62.47$\pm$3.32 & 79.39$\pm$2.85 & 76.40$\pm$2.52 & 66.50$\pm$4.73 & 64.18$\pm$8.39 & 66.23$\pm$2.20 & 51.74$\pm$2.10 & 61.10$\pm$6.06 & 67.90$\pm$3.09 & 82.57$\pm$2.64 \\
8  & Bulrush              & 3,140 & 1,346& 59,601  &   92.61$\pm$0.05 & 91.77$\pm$2.07 & 94.12$\pm$2.16 & 96.27$\pm$0.71 & 95.88$\pm$1.53 & 92.36$\pm$3.51 & 84.42$\pm$28.00& 92.80$\pm$1.40 & 95.91$\pm$1.30 & 89.93$\pm$2.24 & 92.78$\pm$1.57 & 97.47$\pm$0.66 \\
9  & Gravel road          & 1,540 & 644  & 28,547  &   64.12$\pm$0.75 & 75.76$\pm$6.69 & 77.29$\pm$5.33 & 95.52$\pm$1.41 & 91.41$\pm$3.52 & 71.15$\pm$14.03& 83.53$\pm$11.38& 86.54$\pm$0.80 & 78.99$\pm$0.70 & 79.67$\pm$1.60 & 82.12$\pm$3.69 & 96.66$\pm$1.15 \\
10 & Ligustrum vicaryi    & 87    & 37   & 1,659   &   14.91$\pm$2.90 & 64.53$\pm$10.58& 61.36$\pm$5.88 & 82.13$\pm$7.43 & 81.32$\pm$4.91 & 63.08$\pm$19.08& 72.81$\pm$14.34& 76.25$\pm$4.00 & 66.37$\pm$3.90 & 75.87$\pm$4.48 & 58.12$\pm$10.52& 90.44$\pm$3.81 \\
11 & Coniferous pine      & 1,040 & 446  & 19,750  &   1.78$\pm$1.65  & 35.20$\pm$3.27 & 36.30$\pm$2.53 & 61.46$\pm$2.10 & 56.15$\pm$4.42 & 35.72$\pm$5.09 & 39.85$\pm$4.04 & 40.92$\pm$2.60 & 36.70$\pm$2.50 & 35.48$\pm$5.87 & 39.90$\pm$3.69 & 57.25$\pm$2.79 \\
12 & Spiraea              & 36    & 16   & 697     &   6.43$\pm$3.29  & 61.49$\pm$4.97 & 61.65$\pm$10.21& 76.64$\pm$17.60& 83.85$\pm$7.75 & 39.57$\pm$15.73& 17.40$\pm$11.36& 74.03$\pm$2.20 & 73.74$\pm$2.10 & 35.08$\pm$4.67 & 49.84$\pm$8.30 & 87.48$\pm$11.59 \\
13 & Bare soil            & 83    & 35   & 1,568   &   7.00$\pm$0.82  & 95.00$\pm$3.76 & 85.12$\pm$10.28& 98.34$\pm$1.93 & 90.26$\pm$21.68& 85.62$\pm$16.48& 83.43$\pm$29.24& 70.54$\pm$2.40 & 92.86$\pm$2.30 & 94.28$\pm$4.48 & 87.77$\pm$8.51 & 99.16$\pm$0.86 \\
14 & Buxus sinica         & 43    & 19   & 824     &   0.00$\pm$0.00  & 64.75$\pm$5.89 & 66.35$\pm$8.62 & 65.85$\pm$8.51 & 61.35$\pm$19.83& 52.44$\pm$23.61& 50.08$\pm$26.14& 54.73$\pm$8.40 & 60.07$\pm$8.20 & 50.82$\pm$6.02 & 69.49$\pm$6.45 & 59.68$\pm$11.81 \\
15 & Photinia serrulata   & 685   & 294  & 13,041  &   65.24$\pm$0.18 & 65.37$\pm$2.79 & 66.62$\pm$5.97 & 82.85$\pm$2.36 & 75.83$\pm$3.78 & 72.93$\pm$4.53 & 73.78$\pm$3.98 & 69.45$\pm$2.80 & 65.95$\pm$2.70 & 74.44$\pm$4.07 & 65.99$\pm$2.77 & 84.67$\pm$4.89 \\
16 & Populus              & 6,904 & 2,959& 131,041 &   93.69$\pm$0.17 & 80.08$\pm$1.64 & 79.84$\pm$2.50 & 78.79$\pm$2.05 & 78.23$\pm$2.65 & 81.64$\pm$3.86 & 78.12$\pm$2.83 & 84.67$\pm$2.00 & 82.82$\pm$1.90 & 81.55$\pm$3.44 & 81.17$\pm$2.55 & 82.00$\pm$2.56 \\
17 & Ulmus pumila L       & 1,489 & 638  & 7,675   &   66.55$\pm$0.41 & 73.76$\pm$3.42 & 79.06$\pm$3.35 & 80.15$\pm$3.31 & 78.94$\pm$1.42 & 80.79$\pm$3.65 & 75.80$\pm$2.91 & 81.32$\pm$3.50 & 72.83$\pm$3.40 & 77.45$\pm$3.04 & 76.52$\pm$1.21 & 85.07$\pm$2.69 \\
18 & Seawater             & 2,927 & 1,255& 38,093  &   96.88$\pm$0.13 & 88.72$\pm$3.13 & 95.20$\pm$1.73 & 95.52$\pm$2.45 & 95.12$\pm$2.32 & 88.63$\pm$14.98& 92.42$\pm$3.27 & 96.39$\pm$3.70 & 91.41$\pm$3.60 & 94.64$\pm$1.03 & 86.55$\pm$2.74 & 94.46$\pm$2.83 \\

\midrule
\multicolumn{5}{c}{OA (\%)} &
80.09$\pm$0.05 & 80.16$\pm$1.03 & 80.17$\pm$1.08 & 86.77$\pm$0.44 & 85.33$\pm$0.84 & 78.84$\pm$2.47 & 79.48$\pm$4.46 & 84.40$\pm$0.80 & 81.30$\pm$0.75 & 80.89$\pm$1.29 & 81.26$\pm$0.53 & 88.58$\pm$0.44 \\

\multicolumn{5}{c}{AA (\%)} &
54.00$\pm$0.22 & 73.90$\pm$1.09 & 73.87$\pm$1.18 & 84.80$\pm$1.12 & 82.77$\pm$2.40 & 70.64$\pm$2.29 & 71.54$\pm$5.89 & 76.68$\pm$1.20 & 74.08$\pm$1.10 & 73.48$\pm$0.55 & 74.66$\pm$1.35 & 86.94$\pm$1.84 \\

\multicolumn{5}{c}{Kappa $\times$ 100} &
77.00$\pm$0.06 & 77.45$\pm$1.16 & 77.48$\pm$1.19 & 85.06$\pm$0.48 & 83.41$\pm$0.92 & 75.97$\pm$2.77 & 76.78$\pm$5.03 & 82.23$\pm$0.90 & 76.68$\pm$0.85 & 78.27$\pm$1.41 & 78.72$\pm$0.54 & 87.07$\pm$0.47 \\
\bottomrule
\end{tabular}%
}
\end{sidewaystable}